\documentclass[10pt,twocolumn,letterpaper]{article}

\usepackage{cvpr}              

\usepackage{graphicx}
\usepackage{amsmath}
\usepackage{amssymb}
\usepackage{booktabs}
\usepackage{multirow}

\usepackage[pagebackref,breaklinks,colorlinks]{hyperref}

\usepackage[capitalize]{cleveref}
\crefname{section}{Sec.}{Secs.}
\Crefname{section}{Section}{Sections}
\Crefname{table}{Table}{Tables}
\crefname{table}{Tab.}{Tabs.}

\def\cvprPaperID{9801} 
\def\confName{CVPR}
\def\confYear{2022}

\usepackage{booktabs}                                   
\usepackage{multirow}                                   
\usepackage{tablefootnote}                              
\usepackage[symbol]{footmisc}                           
\usepackage{amsmath,amssymb}                            
\usepackage{xcolor}                                     
\usepackage{enumitem}                                   

\newcommand{\eat}[1]{}                                  

\def\wada{W-AdaIN}
\def\rdfg{RDF-GAN}

\begin{document}

\title{RGB-Depth Fusion GAN for Indoor Depth Completion}

\author{Haowen Wang\,\textsuperscript{\rm 1}, Mingyuan Wang\,\textsuperscript{\rm 1},
Zhengping Che\,\textsuperscript{\rm 2}, Zhiyuan Xu\,\textsuperscript{\rm 2},\\
Xiuquan Qiao\,\textsuperscript{\rm 1}\thanks{Corresponding authors.}~, Mengshi Qi\,\textsuperscript{\rm 3},
Feifei Feng\,\textsuperscript{\rm 2}, Jian Tang\,\textsuperscript{\rm 2}\footnotemark[1] \\
\textsuperscript{\rm 1}\,State Key Laboratory of Networking and Switching Technology, Beijing University of Posts and\\ Telecommunications\\
\textsuperscript{\rm 2}\,AI Innovation Center, Midea Group\\
\textsuperscript{\rm 3}\,School of Computer Science, Beijing University of Posts and Telecommunications\\
{\tt\small
\{hw.wang,wmingyuan,qiaoxq,qms\}@bupt.edu.cn\ 
\{chezp,xuzy70,feifei.feng,tangjian22\}@midea.com
}
}

\maketitle

\begin{abstract}
  The raw depth image captured by the indoor depth sensor usually has an extensive range of missing depth values due to inherent limitations such as the inability to perceive transparent objects and limited distance range.
The incomplete depth map burdens many downstream vision tasks, and a rising number of depth completion methods have been proposed to alleviate this issue.
While most existing methods can generate accurate dense depth maps from sparse and uniformly sampled depth maps, they are not suitable for complementing the large contiguous regions of missing depth values, which is common and critical.
In this paper, we design a novel {two-branch} {end-to-end} fusion network, which takes a pair of RGB and incomplete depth images as input to predict a dense and completed depth map.
The first branch employs an encoder-decoder structure to regress the local dense depth values from the raw depth map, with the help of local guidance information extracted from the RGB image.
In the other branch, we propose an RGB-depth fusion GAN to transfer the RGB image to the fine-grained textured depth map.
We adopt adaptive fusion modules named {\wada} to propagate the features across the two branches, and we append a confidence fusion head to fuse the two outputs of the branches for the final depth map.
Extensive experiments on \mbox{NYU-Depth V2} and \mbox{SUN RGB-D} demonstrate that our proposed method clearly improves the depth completion performance, especially in a more realistic setting of indoor environments with the help of the pseudo depth map.

\end{abstract}


\section{Introduction}
\begin{figure}[t]
\centering
\includegraphics[width=0.99\linewidth]{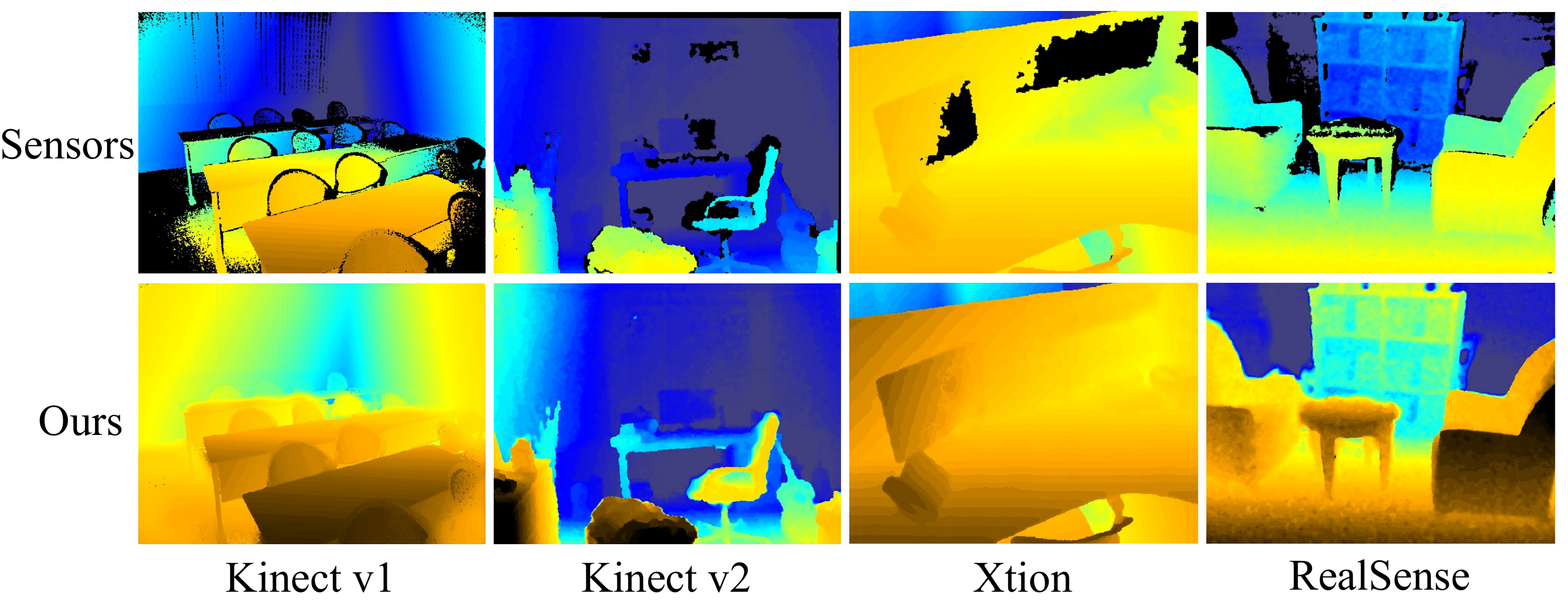}
\caption{
Showcases of the raw depth maps (top) collected by sensors from the {SUN RGB-D} dataset~\cite{song2015sun} and the corresponding depth completion results (bottom) of our method.
}
\label{fig:sunrgb}
\end{figure}

Nowadays, depth sensors have been widely used to provide reliable 3D spatial information in a variety of applications, such as augmented reality, indoor navigation, and 3D reconstruction tasks~\cite{Fu_2018_CVPR,li2019vision,zhao2020pointar}.
However, most existing commercial depth sensors (\textit{e.g.,} Kinect~\cite{kinect}, RealSense~\cite{Keselman_2017_CVPR_Workshops}, and Xtion~\cite{xtion}) for indoor spatial perception are not powerful enough to generate a precise and lossless depth map, as shown in the top row of Fig.~\ref{fig:sunrgb}.
These sensors often produce many hole regions with invalid depth pixels due to transparent, shining, and dark surfaces as well as too close or too far edges,
and these holes significantly affect the performance of downstream tasks on the depth maps (a.k.a., depth images).
To address the issue from imperfect depth maps, there have been a lot of approaches to reconstruct the whole depth map from the raw depth map, called \textit{depth completion}.
As RGB images provide rich color and texture information compared with depth maps, the aligned RGB image is commonly used to guide the depth completion of a depth map.
To be more specific, the depth completion task is usually conducted as using a pair of raw depth and RGB images captured by one depth sensor to complete and refine the depth values. 

Recent studies have produced significant progress in depth completion tasks with convolutional neural networks~(CNNs)~\cite{cheng2018depth,ma2018sparse,qiu2019deeplidar,huang2019indoor,lee2019depth,park2020non}.
Ma and Karaman~\cite{ma2018sparse} introduced an encoder-decoder network to directly regress the dense depth map from a sparse depth map and an RGB image.
The method has shown great progress compared to conventional algorithms~\cite{liu2014discrete,yang2014stereo,saxena2005learning},
but its predicted dense depth maps are often too blurry.
To further generate a more refined completed depth map, lots of works have recently arisen, which can be divided into two groups with different optimization methods.
The first group of works~\cite{liu2017learning,cheng2018depth, park2020non} learn affinities for relative pixels and iteratively refine depth predictions.
These methods highly rely on the accuracy of the raw global depth map and suffer the inference inefficiency.
Other works~\cite{huang2019indoor,lee2019depth,qiu2019deeplidar,lee2021depth} analyze the geometric characteristic and adjust the feature network structure accordingly, for instance, by estimating the surface normal or projecting depth into discrete planes.
These methods require depth completeness without missing regions, and the model parameters may not be efficiently generalized to different scenes.
In any case, the RGB image is merely used as superficial guidance or auxiliary information, and few methods deeply consider the textural and contextual information.
At this point, the depth completion task is more or less degraded to a monocular depth estimation task that is conceptually simple but practically difficult.

More remarkably, most of the above methods~\cite{ma2018sparse,cheng2018depth,lee2021depth} uniformly randomly sample a certain number of valid pixels from the dense depth image $\mathbf{d}_{raw}$ and $\mathbf{d}_{gt}$ to mimic the sparse depth map $\mathbf{d}^*$ for training and evaluation, respectively.
Such sampling strategy is credible in some scenes, such as the outdoor range-view depth map generated by LiDAR.
However, the sampled patterns are quite different from the real missing patterns, such as the large missing regions and semantic missing patterns shown in Fig.~\ref{fig:sunrgb}, in indoor depth maps.
Therefore, though existing methods are shown to be effective for completing uniformly sparse depth maps, it remains unverified whether they perform well enough for indoor depth completion.\footnote[2]{Please refer to Section~1 of the supplementary for more discussions.}

To solve these problems, we propose a novel \mbox{two-branch} end-to-end network to generate a completed dense depth map for indoor environments.
Inspired by generative adversarial networks~(GANs)~\cite{mirza2014conditional,isola2017image,karras2019style,ma2019fusiongan},
we introduce the \textit{RGB-depth Fusion GAN}~(\rdfg) for fusing an RGB image and a depth map.
\mbox{RDF-GAN} maps a conditional RGB image from the RGB domain to a dense depth map from the depth domain through the latent spatial vector generated by the incomplete depth map.
We further design a \textit{constraint network} to restrict the depth values of the fused map, with the help of \textit{weighted-adaptive instance normalization}~(\wada) modules and a \textit{local guidance module}.
Afterwards, a \textit{confidence fusion head} concludes the final depth map completion.

In addition, we propose an exploitation technique, which samples raw depth images to produce pseudo depth maps for training.
According to the characteristic of the indoor depth missing, we utilize the RGB images and semantic labels to produce masking regions for raw depth maps, which is more realistic than the simple uniform sampling.
Experiments show that the model learning from pseudo depth maps can more effectively fill in large missing regions for raw depth images captured indoors.

Our main contributions are summarized as the following:
\begin{itemize} [itemsep=1pt,topsep=1pt,parsep=1pt,leftmargin=10pt] 
    \item We propose a novel end-to-end GAN-based network, which effectively fuses a raw depth map and an RGB image to reproduce a reasonable dense depth map.
    \item We design and utilize the pseudo depth maps, which are in line with the raw depth missing distribution in indoor scenarios. Training with pseudo depth maps significantly improves the model's depth completion performance, especially in more realistic settings of indoor environments.
    \item Our proposed method achieves the state-of-the-art performance on {NYU-Depth V2} and {SUN RGB-D} for depth completion and proves its effectiveness in improving downstream task performance such as object detection.
\end{itemize}

\begin{figure*}[t!]
  \centering
  \includegraphics[width=1\linewidth]{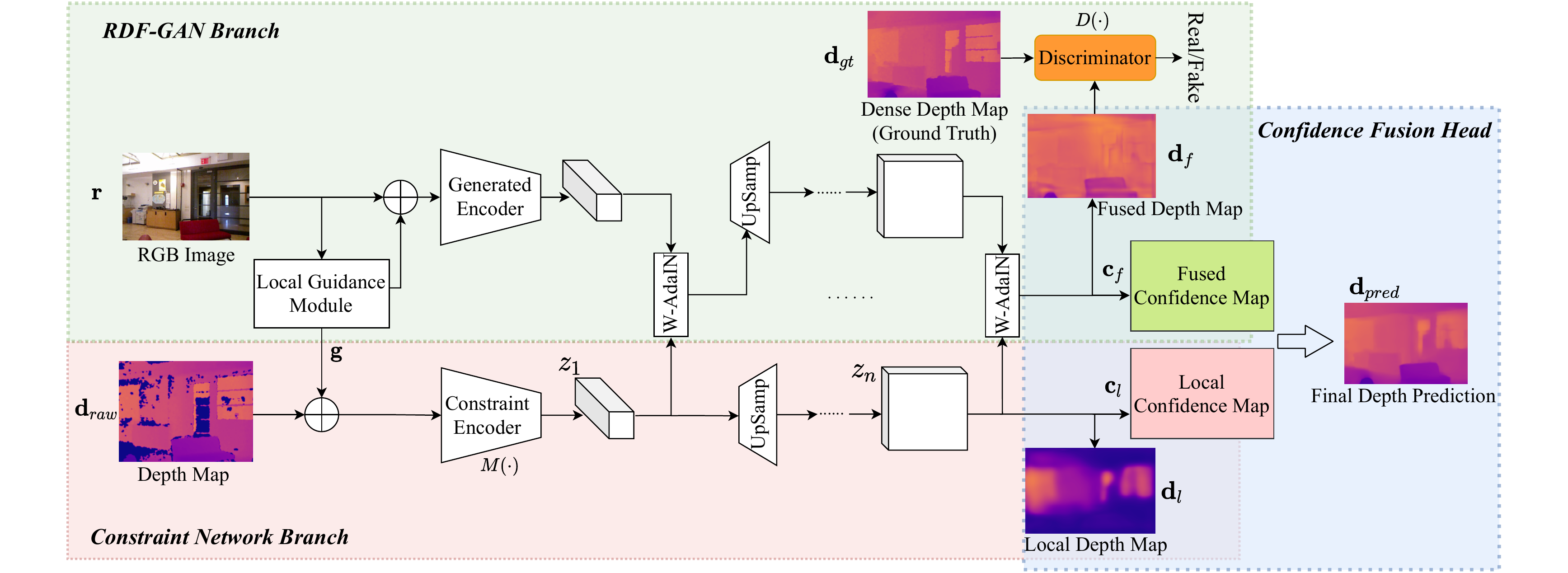}
  \caption{The overview of the proposed end-to-end depth completion method. The architecture consists of two main components: the \mbox{\rdfg} branch and the constraint network branch. The generator of {\rdfg} generates a fused depth map, which is distinguished as real/fake by the discriminator. The local guidance module and {\wada} share the features across stages. The confidence fusion head merges the fused depth map and local depth map generated by the two branches to produce the final prediction.}
  \label{fig:architect}
\end{figure*}

\section{Related Work}
\textbf{Depth Completion.}
Recent works have extensively applied deep neural networks for depth estimation and completion tasks with remarkable improvements.
Ma and Karaman~\cite{ma2018sparse} used an encoder-decoder structure with CNNs to predict the full-resolution depth image directly from a set of depth samples and RGB images.
On this basis, some methods incorporating additional output branches to assist in the generation of depth maps have been proposed.
Qiu~\etal~\cite{qiu2019deeplidar} produced dense depth using the surface normal as the intermediate representation.
Huang~\etal~\cite{huang2019indoor} applied the boundary consistency to solve the issue of vague structures.
Lee~\etal~\cite{lee2021depth} introduced the Plane-Residual representation to interpret depth information and factorized the depth regression problem into a combination of discrete depth plane classification and plane-by-plane residual regression.
Zhang~\etal~\cite{depthcompletion_gan} uses GANs to solve both semantic segmentation and depth completion tasks in outdoor scenarios.
Cheng~\etal~\cite{cheng2018depth} proposed the convolutional spatial propagation network~(CSPN) and generated the long-range context through a recurrent operation to lessen the burden of directly regressing the absolute depth information.
Park~\etal~\cite{park2020non} improved CSPN by non-local spatial and global propagations.
These methods prove that the encoder-decoder network can effectively perform depth completion and obtain a more refined depth map through additional optimization. 
In this work, we extend the encoder-decoder structure to build our depth completion model.

\textbf{RGB-D Fusion.}
The fusion of both RGB and depth data (a.k.a., the RGB-D fusion) is essential in many tasks such as semantic segmentation and depth completion.
While most existing methods~\cite{maire2016affinity,ma2018sparse} only concatenate aligned pixels from RGB and depth features, more effective and advanced RGB-D fusions have been proposed recently.
Cheng~\etal~\cite{cheng2017locality} designed a gated fusion layer to learn the different weights of each modality in different scenes.
Park~\etal~\cite{park2017rdfnet} fused multi-level RGB-D features in a very deep network through residual learning.
Du~\etal~\cite{du2019translate} proposed a novel cross-modal translate network to represent the complementary information and enhance the discrimination of extracted features.
In this work, we design the two-branch structure and the W-AdaIN modules to better capture and fuse RGB and depth features.

\textbf{Generative Adversarial Networks.}
Generative adversarial networks~(GANs) have achieved great success in a variety of image generation tasks such as image-style transfer, realistic image generation, and image synthesis.
Mirza~\etal~\cite{mirza2014conditional} proposed the conditional GAN to direct the data generation process by combining the additional information as a condition.
Karras~\etal~\cite{karras2019style} introduced a style-based GAN to embed the latent code into a latent space to affect the variations of generated images.
Ma~\etal~\cite{ma2019fusiongan} proposed a GAN for infrared and visible images.
In this work, we use a GAN-based structure fusing RGB images and depth maps to generate dense depth maps with fine-grained textures.

\section{Method}
In this section, we describe our end-to-end depth completion method,
as shown in Fig.~\ref{fig:architect}.
The proposed model takes a raw (noisy and possibly incomplete) depth map and its corresponding RGB image as the input, and outputs the completed and refined dense depth map estimation.
The model mainly consists of two branches: a constraint network branch (Section~{\ref{sec:local_dc}}) and an RGB-depth Fusion GAN~({\rdfg}) branch (Section~{\ref{sec:gan_dc}}).
The constraint network and {\rdfg} take a depth map and an RGB image as the input, respectively, and produce their depth completion results.
To fuse the representations between the two branches,
a local guidance module and a series of intermediate fusion modules called {\wada} (Section~{\ref{sec:connect_module}}) are deployed at different stages of the model.
Finally, a confidence fusion head (Section~{\ref{sec:confidence_fusion}}) combines the outputs of the two channels and provides more reliable and robust depth completion results.
Moreover, we introduce the training strategy with pseudo depth maps (Section~\ref{sec:pseudo depth map}) and describe the overall loss function for training (Section~\ref{sec:loss_function}).

\subsection{Constraint Network Branch}
\label{sec:local_dc}
\begin{figure}[t]
\centering
\includegraphics[width=0.97\linewidth]{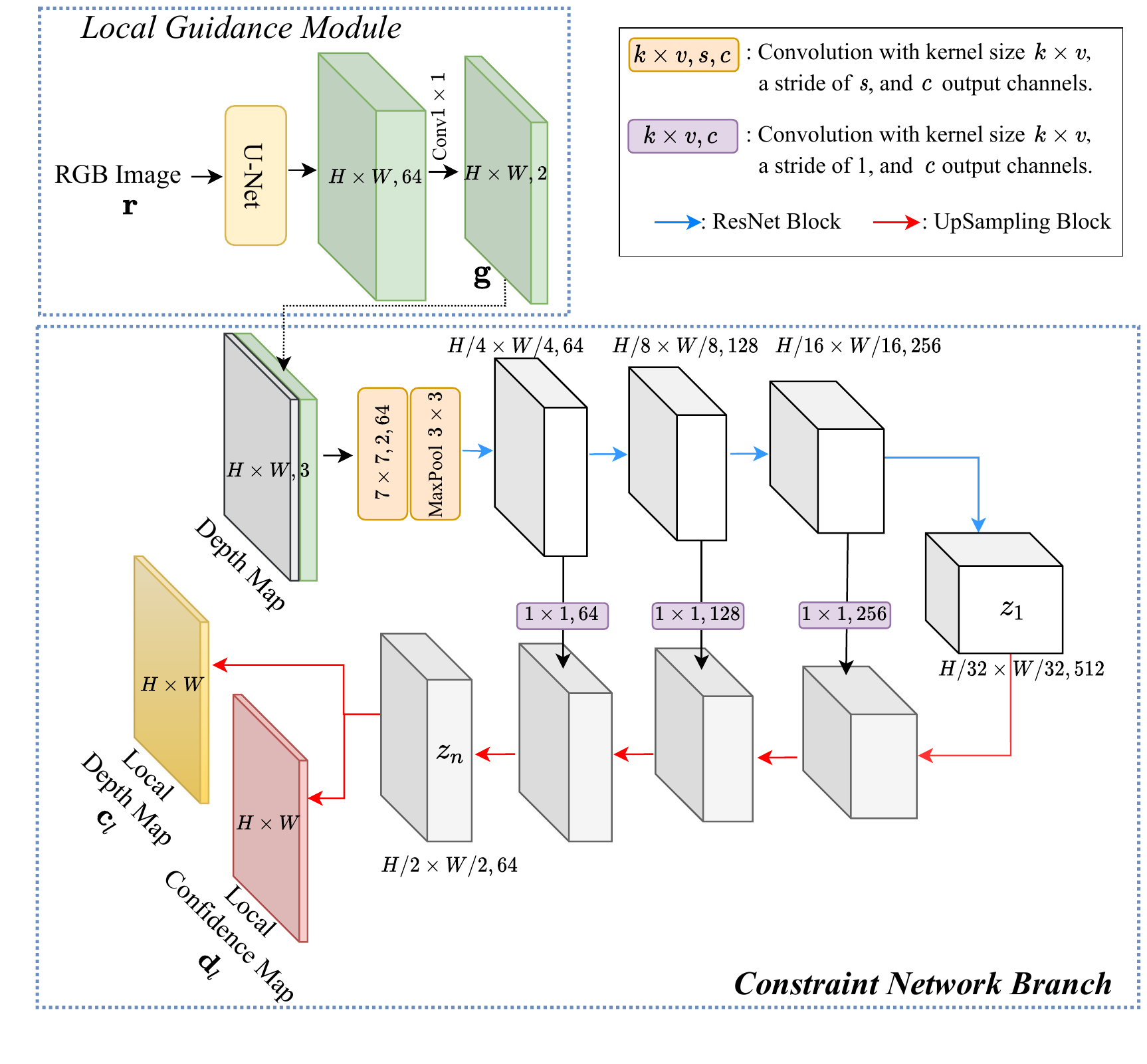}
\caption{An Illustration of the constraint network.}
\label{fig:local_dc}
\end{figure}

The first branch is composed of a constraint network, which reproduces a local full-resolution depth map and a confidence map through a convolutional encoder-decoder structure.
The encoder-decoder structure is based on ResNet-18~\cite{he2016deep} and pre-trained on the ImageNet dataset~\cite{deng2009imagenet}.
As illustrated in Fig.~\ref{fig:local_dc} and the bottom-left part of Fig.~\ref{fig:architect},
given the raw depth image $\mathbf{d}_{raw} \in \mathbb{R}^{H \times W \times 1}$ and the RGB image $\mathbf{r}$, the network outputs a dense local depth map $\mathbf{d}_{l} \in \mathbb{R}^{H \times W \times 1}$ and a local confidence map $\mathbf{c}_{l} \in \mathbb{R}^{H \times W \times 1}$.

The input of this branch is a concatenation of the one-channel raw depth image $\mathbf{d}_{raw}$ and the two-channel local guidance map $\mathbf{g}$ from the RGB image.
Given this input, the encoder downsamples the feature size to $\frac{H}{32} \times \frac{W}{32}$ and expands the feature dimension to 512.
The encoder~$M(\cdot)$ learns the mapping from the depth map to the depth latent space $\mathbf{\it z}$ as the fused depth feature information for {\rdfg}.
The decoding stage applies a set of upsampling blocks to increase the feature resolution with skip connection from the encoder.
The output of the decoder is a local depth map and its corresponding local confidence map.

\begin{figure}[t]
\centering
\includegraphics[width=.75\linewidth]{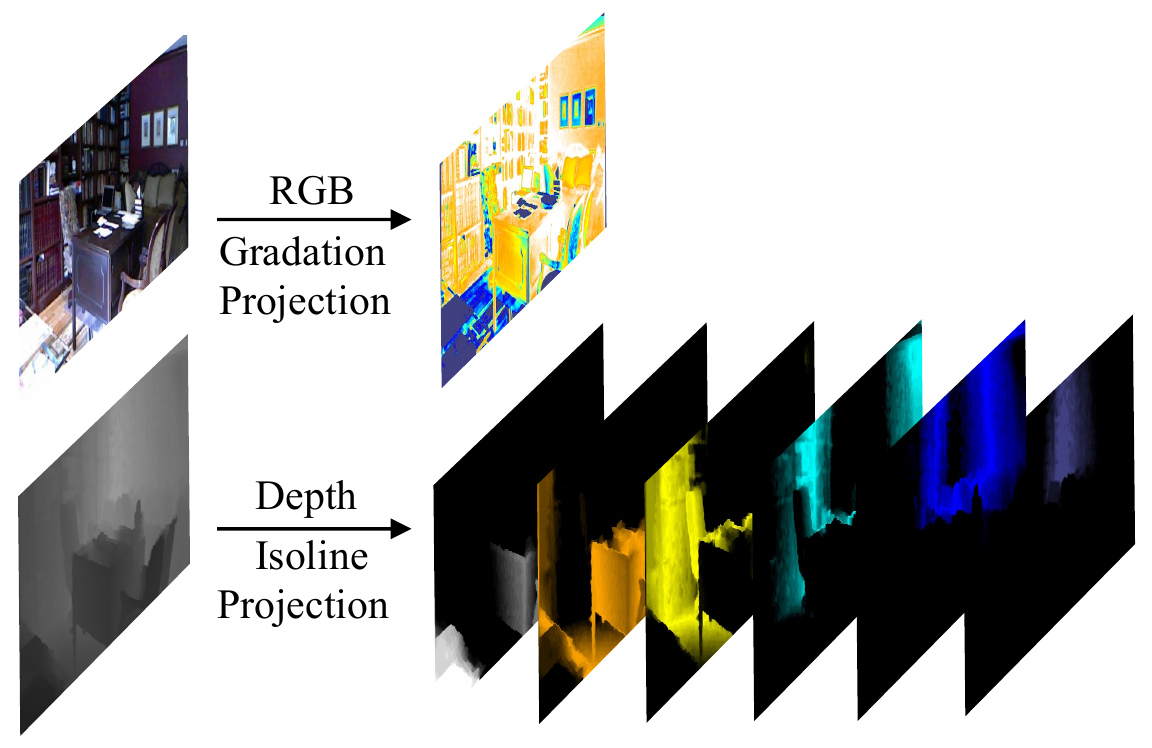}
\caption{An example of projecting an RGB image to color gradations (top) and projecting a depth map to different depth planes.}
\label{fig:projection}
\end{figure}

\subsection{RDF-GAN Branch}
\label{sec:gan_dc}
To generate the fine-grained textured and dense depth map, we propose the second branch in our model, which is a GAN-based structure for RGB and depth image fusion.
Different from most existing fusion methods that directly concatenate inputs from different domains, our fusion model, named as {\rdfg}, is inspired by the conditional and style GANs~\cite{mirza2014conditional,karras2019style}.
As illustrated in the top-left part of Fig.~\ref{fig:architect},
we use the depth latent vector mapping from incomplete depth image as the input and the RGB image as the condition to generate a dense fused depth prediction and a fused confidence map, and use a discriminator to distinguish the real~(ground truth) depth images from generated ones.
The generator $G(\cdot)$ has a similar structure to the constraint network.
Given the crosseponding RGB image $\mathbf{r}$ as the condition, the generator $G(\cdot)$ with the depth latent vector $z$ generates a fused dense depth map $\mathbf{d}_f$ and a fused confidence map $\mathbf{c}_f \in \mathbb{R}^{H \times W \times 1}$ for the scene.
The latent vector $z$ propagates the depth information to the RGB image using the proposed {\wada} described in Section~\ref{sec:connect_module}.
We distinguish the fused depth map $\mathbf{d}_f$ and the real depth image $\mathbf{d}_{gt}$ by the discriminator ${D(\cdot)}$, whose structure is based on PatchGAN~\cite{isola2017image}.
We adopt the objective function of WGAN~\cite{arjovsky2017wasserstein} for training {\rdfg}.
To be more specific, the {\rdfg} loss includes the discriminator loss ${L}_{D}$ and the generator loss ${L}_{G}$:
\begin{equation}
\begin{aligned}
{L}_{D}= \mathop{\mathbb{E}}_{\mathbf{d}_{\text {raw}} \sim \mathcal{D}_{\text {raw}}}
\left[D(G\left(M({\mathbf{d}_{\text {raw }}})\right)| \mathbf{r}\right]
- \mathop{\mathbb{E}}_{\mathbf{d}_\text{gt}\sim \mathcal{D}_{\text {gt}}}\left[D(\mathbf{d}_{\text {gt }}|\mathbf{r}\right)],
\label{equption:gan_dloss}
\end{aligned}
\end{equation}
\begin{equation}
\begin{aligned}
{L}_{G}= {\lambda}_g L_{1}(G(M({\mathbf{d}_{\text{raw}}})) - \mathop{\mathbb{E}}_{\mathbf{d}_{\text {raw}} \sim \mathcal{D}_{\text {raw}}}\left[D(G\left(M({\mathbf{d}_{\text {raw }}})\right) |\mathbf{r}\right],
\label{equption:gan_gloss}
\end{aligned}
\end{equation}
where $d_{\text{raw}}$ and ${d}_{gt}$ are the raw and ground-truth depth images drawn from the domains $\mathcal{D}_{\text{raw}}$ and $\mathcal{D}_{\text {gt}}$, respectively.

\subsection{Feature Fusion Modules}
\label{sec:connect_module}
To allow the feature information to be shared across all stages of the two branches, we design the local guidance module and {\wada} and apply them in the network.

\textbf{Local Guidance Module.}
We adopt U-Net~\cite{ronneberger2015u} as a feature extractor to produce a local guidance map $\mathbf{g} \in \mathbb{R}^{H \times W \times 2}$ from an RGB image $\mathbf{r} \in \mathbb{R}^{H \times W \times 3}$.
The first and the second channels of the local guidance map represent the foreground probability and semantic features, respectively.
Therefore, the local guidance module can guide the constraint network to focus on local depth correlations.

\textbf{\wada.}
As shown in Fig.~\ref{fig:projection}, we project depth pixels of the depth map into multiple discretized depth planes, according to the distance between the depth pixels and a pre-defined set of discrete depth values. Local regions are easier to be classified into the same depth plane because they have similar depth values.
We also find that similar color gradations in a local region usually have similar depth values.
Hence, we propose a \mbox{\wada} module for fusing the features of RGB and depth images.
It is extended from AdaIN~\cite{karras2019style} and is defined as:
\begin{equation}
    \operatorname{W-AdaIN}(z,f_r)    =A \cdot y_s  \cdot \left(\frac{f_r-\mu(f_r)}{\sigma(f_r)}\right)+B \cdot y_b,
\label{equption:wada}
\end{equation}
where
$f_r$ is the feature map of RGB image;
$A = \operatorname{Attention}(z)$ and $B =  \operatorname{Attention}(f_r)$ are the weight matrices that are generated by the self-attention mechanism~\cite{vaswani2017attention} on $z$ and $f_r$, respectively;
$y_s$ and $y_b$ are the spatial scaling and bias factors obtained by affine transformations~\cite{karras2019style} with the latent matrix $z$;
$\mu(\cdot)$ and $\sigma(\cdot)$ are the mean and variance, respectively.
By its design,  $A$ assigns similar weight values to the regions with similar depth values.
Similarly, $B$ smoothes the depth blocks by assigning similar weight values of the local similar color gradations.

\subsection{Confidence Fusion Head}
\label{sec:confidence_fusion}
In our framework, either branch has its role.
The RDF-GAN branch estimates the missing depth based on the textural features of the RGB image but may produce obvious outliers, i.e., estimation deviations from the raw depth values.
The constraint network branch, with an encoder-decoder structure, generates a locally accurate depth map by relying more on valid raw depth information.
Hence, we introduce the confidence maps~\cite{van2019sparse} to integrate the depth maps from two branches by a confidence fusion head, which is shown in the right of Fig.~\ref{fig:architect}.
We introduce the confidence maps~\cite{van2019sparse} of both branches to assign more attention to reliable depth prediction regions through the learned confidence.
In general, the local depth map obtains higher confidences in regions whose raw depth values are more accurate, while the fused depth map has higher confidences in large missing and noisy regions.
The sum of the two depth maps weighted by the corresponding confidence maps is the final depth prediction, which is formulated as:
\begin{equation}
{d}_{pred }(i, j)=\frac{e^{c_{l}(i, j)} \cdot {d}_{ {l}}(i, j)+e^{c_{f}(i, j)} \cdot {d}_{{f}}(i, j)}{e^{c_{l}(i, j)}+e^{c_{f}(i, j)}}.
\end{equation}

\begin{figure}[t]
\centering
\includegraphics[width=0.99\linewidth]{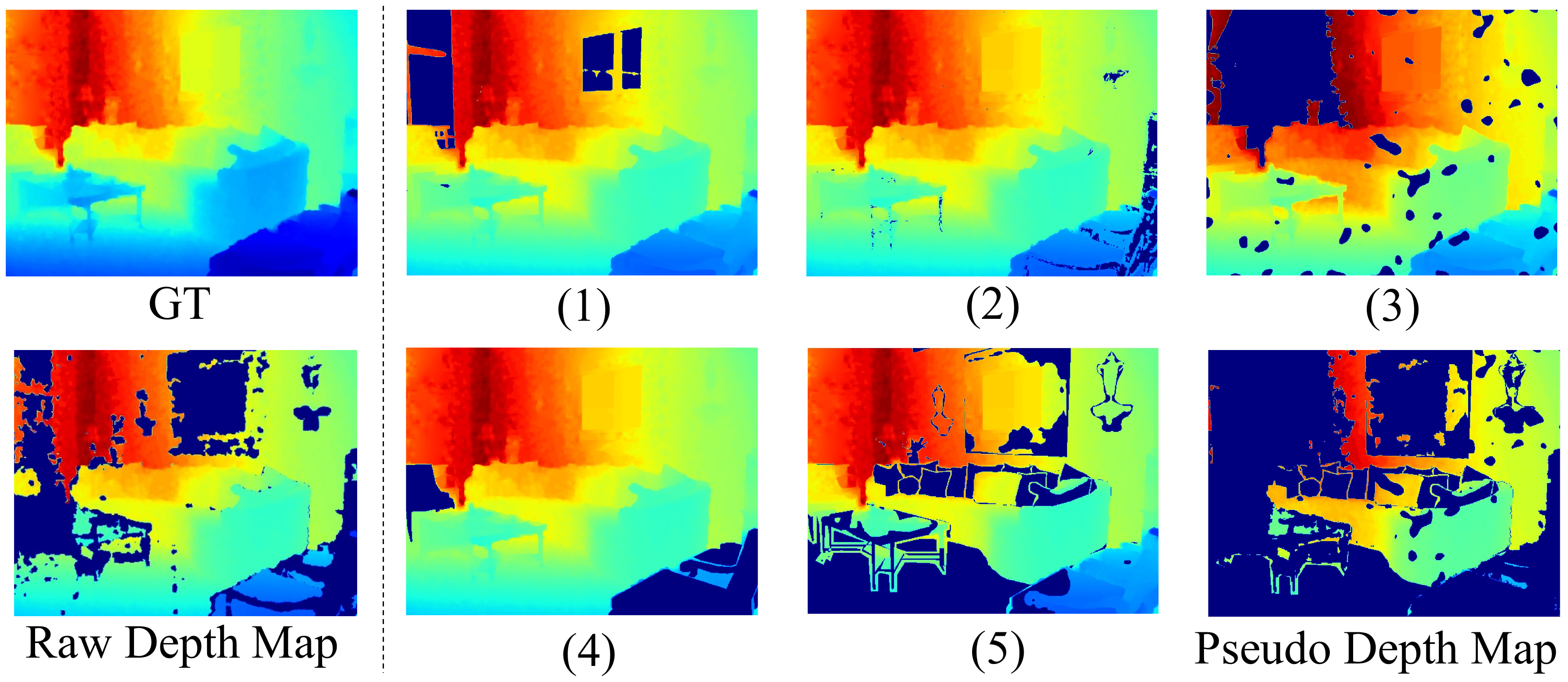}
\caption{Visualizations of the proposed pseudo depth map and five sampling methods.
GT represents the reconstructed (ground-truth) depth map.
The shown pseudo depth map is generated from the raw depth image by applying all sampling methods together.}
\label{fig:pseudo}
\end{figure}

\subsection{Pseudo Depth Map for Training}
\label{sec:pseudo depth map}
Most existing depth completion methods are trained and evaluated with the random sparse sampling method~\cite{ma2018sparse,park2020non,lee2021depth}.
The sampled depth map mimics outdoor depth well, but its depth distribution and missing patterns are quite different from the real indoor depth completion scene.
The randomly downsampled depth pixels cover almost all areas of the scene, while the missing depth pixels in indoor environments usually form continuous areas.
Hence, we propose a set of synthetic methods to produce depth maps for model training, which rely on RGB images and semantic masks to map the raw depth image to reasonable incomplete (pseudo) depth maps.
Pseudo depth map mimics the depth missing patterns and is more like real raw depth data than the randomly sampled depth maps.

We design five methods to obtain the pseudo depth map:
\begin{enumerate} [itemsep=1pt,topsep=1pt,parsep=1pt,leftmargin=20pt,label=(\arabic*)] 
    \item \textit{Highlight masking}. We segment the regions of probably specular highlights~\cite{arnold2010automatic} in RGB images and mask them in raw depth maps.
    \item \textit{Black masking}. We randomly mask the depth pixels whose RGB values are all in [0, 5] (i.e., dark pixels).
    \item \textit{Graph-based segmentation masking}. We mask the probably noisy pixels of depth maps obtained by graph-based segmentations~\cite{felzenszwalb2004efficient} on RGB images.
    \item \textit{Semantic masking}. As depth values for objects with some particular materials are usually missing, we mask one or two objects randomly by their semantic labels and only keep depth pixels on their edges.
    \item \textit{Semantic XOR masking}. We train U-Net~\cite{ronneberger2015u} on 20\% of the training set of RGB images and use the trained model to segment the other RGB images.
        We mask the depth pixels where the segmentation result and ground-truth are different, i.e., conducting the XOR operation on the segmentation results and the ground-truth to obtain the masking.
\end{enumerate}

Finally, we randomly pick and combine the mask from the above five methods to generate the pseudo depth map, mimicking a more plausible missing depth distribution. The pseudo depth maps are used to train a more robust depth completion model for indoor scenarios.
More details can be found in Section~2 of the supplementary.

\subsection{Loss Function}
\label{sec:loss_function}
We use the $L_1$ loss on the local depth map and final prediction.
The overall loss function is defined as:
\begin{equation}
L_{overall}=L_{D} + L_{G}+\lambda_{l} L_{1}(\mathbf{d}_{l})+\lambda_{pred} L_{1}(\mathbf{d}_{pred}),
\end{equation}
where 
$\lambda_{g}$ in Eq.~\ref{equption:gan_gloss}, $\lambda_{l}$, and $\lambda_{pred}$ are weight hyperparameters for different terms in the loss function, which are set to be 0.5, 1, and 10, respectively.

\section{Experiments}
\subsection{Datasets and Metrics}
\label{sec:ex_datasets}
We conducted experiments on two widely-used benchmarks: \mbox{NYU-Depth V2}~\cite{Silberman:ECCV12} and \mbox{SUN RGB-D}~\cite{song2015sun}.

\textbf{NYU-Depth V2.}
The NYU-Depth V2 dataset~\cite{Silberman:ECCV12} contains pairs of RGB and depth images collected from Microsoft Kinect in 464 indoor scenes.
Densely labeled image pairs are split into the training set with 795 images and the test set with 654 images, and each set includes RGB images, raw depth images from sensors, labeled~(reconstructed) depth maps, and segmentation masks.
Following existing methods, we utilized the unlabeled $\sim$50K images for training and the labeled 654 images in the test set for evaluation.
The input images were resized to 320$\times$240 and center-cropped cropped to 304$\times$228.

\textbf{SUN RGB-D.}
The SUN RGB-D dataset~\cite{song2015sun} contains 10,335 RGB-D images captured by four different sensors.
This dataset, with different scenes and sensors, is diverse and helpful to effectively evaluate model generalization.
Besides, its dense semantic annotations and 3D bounding boxes enable the evaluations of more training strategies and downstream tasks.
Following the official split, we used 4,845 images for training and 4,659 for testing in 19 major scene categories.
We used the refined depth map based on multiple frames~\cite{song2015sun} as the ground truths for evaluation.
The input images were resized to 320$\times$240 and randomly cropped to 304$\times$228.

\textbf{Evaluation Metrics.}
We adopted three metrics for the dense depth prediction evaluation: root mean squared error~(RMSE), absolute relative error~(Rel), and $\delta_{i}$, which is the percentage of predicted pixels whose relative error is within a relative threshold~\cite{ma2018sparse}.

\subsection{Comparisons with State-of-the-Art Methods}
\label{sec:compar_sota}
\begin{table}[t]
\centering
\resizebox{1\columnwidth}{!}
    {
    \begin{tabular}{c|c|c|c|ccc}
        \toprule
        Setting & Method & RMSE $\downarrow$ & Rel $\downarrow$ & $\delta_{1.25} \uparrow$ & $\delta_{1.25^{2}} \uparrow$ & $\delta_{1.25^{3}} \uparrow$ \\
        \midrule
        \midrule
        \multirow{6}{*}{$\mathcal{R} \Rightarrow \mathcal{T}$}
        & DC-BCS~\cite{huang2019indoor} & $0.271$ & $0.016$ & $98.1$ & $99.1$ & $99.4$ \\
        & RGB-GU~\cite{van2019sparse} & $0.260$ & $0.017$ & $97.9$ & $99.3$ & $99.7$ \\
        & MS-CHN~\cite{li2020multi} & $0.190$ & $0.018$ & $98.8$ & $99.7$ & $99.9$ \\
        & DM-LRN~\cite{senushkin2020decoder} & $0.205$ & $0.014$ & $98.8$ & $99.6$ & $99.9$ \\
        & NLSPN~\cite{park2020non} & $0.153$ & $0.015$ & $98.6$ & $99.6$ & $99.9$ \\
        & Ours & $\textbf{0.139}$ & $\textbf{0.013}$ & $98.7$ & $99.6$ & $99.9$ \\
        \midrule
        \multirow{4}{*}{$\mathcal{R^*} \Rightarrow \mathcal{T}$} & Sparse2Dense~\cite{ma2018sparse} & $0.335$ & $0.060$ & $94.2$ & $97.1$ & $98.8$ \\
         & CSPN~\cite{cheng2018depth} & $0.500$ & $0.139$ & $85.7$ & $92.9$ & $96.3$ \\
         & NLSPN~\cite{park2020non} & $0.348$ & $\textbf{0.043}$ & $93.0$ & $96.7$ & $98.5$  \\
        & Ours & $\textbf{0.309}$ & $0.053$ & $93.6$ & $97.6$ & $99.0$ \\
        \midrule
        \multirow{8}{*}{$\mathcal{T^*} \Rightarrow \mathcal{T}$} & Sparse2Dense~\cite{ma2018sparse}  & $0.230$ & $0.044$ & $97.1$ & $99.4$ & $99.8$ \\
        & CSPN~\cite{cheng2018depth} & $0.117$ & $0.016$ & $99.2$ & $99.9$ & $100.0$\\
        & 3coeff~\cite{imran2019depth} & $0.131$ & $0.013$ & $97.9$ & $99.3$ & $99.8$ \\
        & DGCG~\cite{lee2019depth}  & $0.225$ & $0.046$ & $97.2$ & $-$ & $-$\\
        & DeepLidar~\cite{qiu2019deeplidar} & $0.115$ & $0.022$ & $99.3$ & $99.9$ & $100.0$ \\
      & NLSPN~\cite{park2020non} & $\textbf{0.092}$ & $\textbf{0.012}$ & $99.6$ & $99.9$ & $100.0$  \\
        & PRR~\cite{lee2021depth} & $0.104$ & $0.014$ & $99.4$ & $99.9$ & $100.0$\\
        & Ours & $0.103$ & $0.016$ & $99.4$ & $99.9$ & $100.0$ \\
        \bottomrule
    \end{tabular}
    }
    \caption{Quantitative results on the NYU-Depth V2 dataset. $\mathcal{R}$ and $\mathcal{T}$ represent the raw and reconstructed depth map, respectively. \\
    $\cdot^*$ represents the random sparse sampling, where Spares2Dense and DGCG in {$\mathcal{T^*} \Rightarrow \mathcal{T}$} use 200 pixels and others use 500 pixels.
    }
    \label{tab:nyu}
\end{table}

\textbf{NYU-Depth V2.}
To draw a comprehensive performance analysis, we set up three different training and evaluation schemes.
In the test, we use three different inputs to predict and reconstruct depth maps $\mathcal{T}$ respectively, which are raw depth maps $\mathcal{R}$, sparse depth maps with randomly sampled 500 valid depth pixels in raw depth map $\mathcal{R^*}$, and sparse depth maps with randomly sampled 500 valid depth pixels in reconstructed depth map $\mathcal{T^*}$. For more descriptions of the schemes, please refer to Section~3 in the supplementary.
The performance comparison of our method and the other state-of-the-art methods on NYU-Depth V2 are shown in Tab.~\ref{tab:nyu}. Given the results, we concluded the following:

\begin{figure*}[t]
\centering
\includegraphics[width=.93\linewidth]{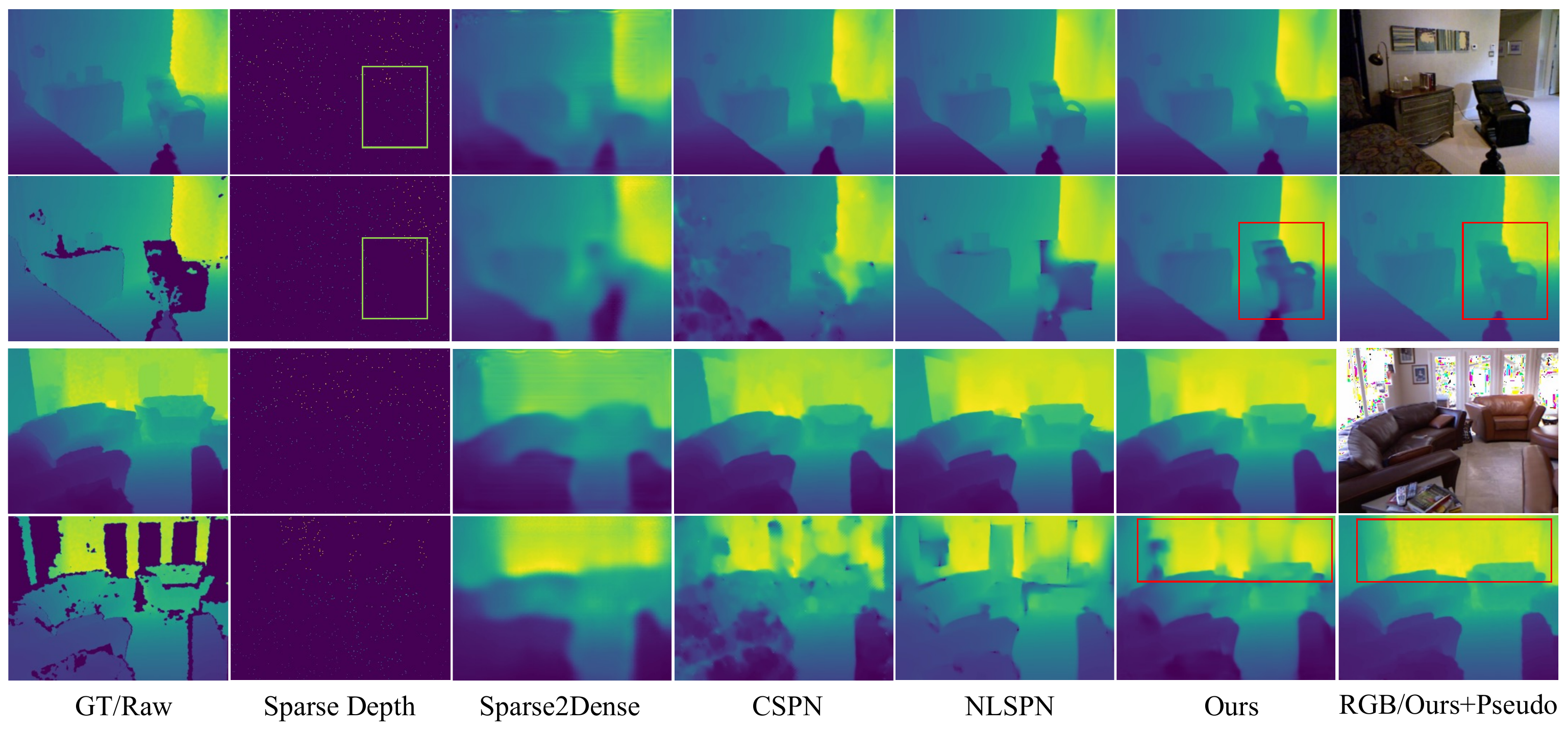}
\caption{Depth completion comparisons of different methods with different training strategies and inputs on NYU-Depth V2.
The first and third rows take sparse samples on reconstructed depth maps as the input ($\mathcal{T^*} \Rightarrow \mathcal{T}$).
The second and fourth rows take sparse samples on raw depth maps as the inputs ($\mathcal{R^*} \Rightarrow \mathcal{T}$).
The last column shows the result of our model trained with pseudo maps ($\mathcal{R} \Rightarrow \mathcal{T}$).}
\label{fig:nyu_raw}
\end{figure*}

\begin{figure*}[t]
\centering
\includegraphics[width=.93\linewidth]{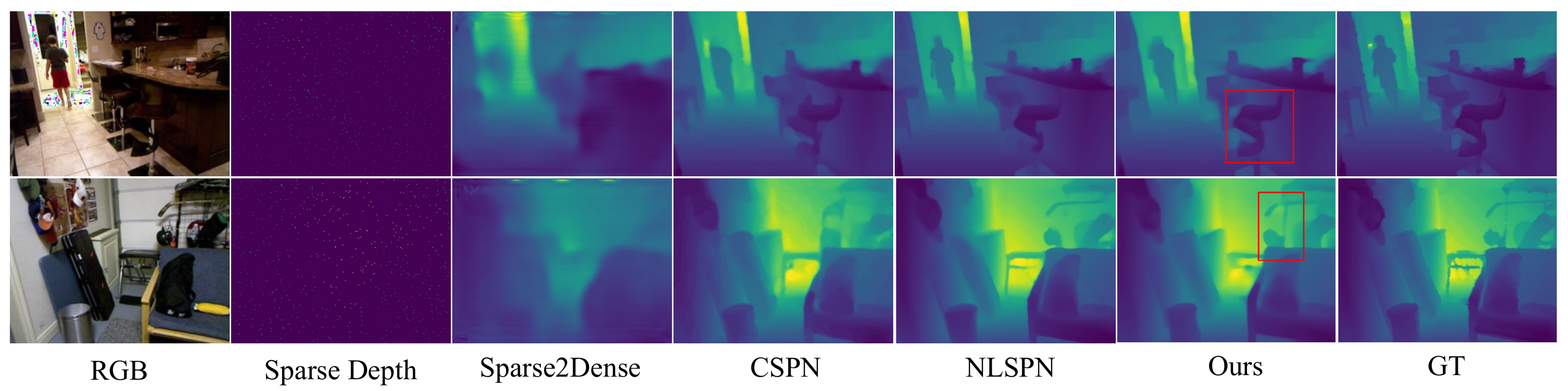}

\caption{
Depth completion comparisons on NYU-Depth V2 with $\mathcal{T^*} \Rightarrow \mathcal{T}$.
Our model recovers more textural details in the red boxes.
}
\label{fig:nyu_compare}
\end{figure*}

\begin{itemize} [itemsep=1pt,topsep=1pt,parsep=1pt,leftmargin=10pt] 
  \item $\mathcal{R} \Rightarrow \mathcal{T}$:
    We used the pseudo depth maps generated in Section~\ref{sec:pseudo depth map} as the input to train the proposed model and NLSPN~\cite{park2020non}.
    Meanwhile, we compared with several baselines~\cite{senushkin2020decoder,huang2019indoor,van2019sparse,li2020multi} that are trained in the synthetic semi-dense sensor data~\cite{senushkin2020decoder}.
    Compared to all the baselines, our proposed method improves significant performance, especially on RMSE and Rel.
    We selected two representative scenes and visualized our prediction results in the last column of Fig.~\ref{fig:nyu_raw}.
    The model trained by pseudo depth maps produced more accurate and textured depth predictions in the missing depth regions.

  \item $\mathcal{R^*} \Rightarrow \mathcal{T}$:
    Following the previous works~\cite{ma2018sparse,cheng2018depth,park2020non,lee2021depth}, we used the RGB image and the sparse depth map with randomly sampled depth pixels of raw depth image as the input for training.
    In the test stage, the input was the same as that for training, and the reconstructed depth map was used as the ground truth.
    We observed that our model outperformed the baseline with big margins on RMSE. The qualitative results were shown in the second and fourth rows of Fig.~\ref{fig:nyu_raw}.
    Our method accurately predicts the contour of the sofa and smooth windows in red boxes compared to other methods.
    This proves that our dense depth predictions are well integrated with the textural information of RGB images by the RDF-GAN branch.

  \item $\mathcal{T^*} \Rightarrow \mathcal{T}$:
    The setting is consistent with most existing works of depth completion~\cite{ma2018sparse,cheng2018depth,park2020non,lee2021depth}.
    Our model without any iteration processing is only lower than the NLSPN~\cite{qiu2019deeplidar} (but ours is 1.5$\times$ faster in inference time than NLSPN). The visualizations shown in the first and third rows of Fig.~\ref{fig:nyu_raw} as well as Fig.~\ref{fig:nyu_compare} further indicate the superiority of our method.

  \item
    As shown in green boxes of Fig.~\ref{fig:nyu_raw}, the downsampled input from the reconstructed depth map ($\mathcal{T^*} \Rightarrow \mathcal{T}$) reveals ground truth depth values, which is unavailable in practice, to the models. This supported the claim that the raw input setting ($\mathcal{R} \Rightarrow \mathcal{T}$) is more practicable for realistic indoor depth completion.
\end{itemize}

\begin{table}[t]
\resizebox{1\columnwidth}{!}
    {
    \begin{tabular}{c|c|c|ccc}
   \toprule
   $\mathcal{R} \Rightarrow \mathcal{T}$  & RMSE $\downarrow$ & Rel $\downarrow$ & $\delta_{1.25} \uparrow$ & $\delta_{1.25^{2}} \uparrow$ & $\delta_{1.25^{3}} \uparrow$ \\
    \midrule
    Sparse2Dense~\cite{ma2018sparse} & $0.329$ & $0.074$ & $93.9$ & $97.0$ & $98.1$\\
    CSPN~\cite{cheng2018depth} & $0.295$ & $0.137$ & $95.6$ & $97.5$ & $98.4$ \\
    DeepLidar~\cite{qiu2019deeplidar} & $0.279$ & $0.061$ & $96.9$ & $98.0$ & $98.4$\\
    NLSPN~\cite{park2020non} & $0.267$ & $0.063$ & $97.3$ & $98.1$ & $98.5$ \\
    Ours & $\textbf{0.255}$ & $\textbf{0.059}$ & $96.9$ & $98.4$ & $99.0$ \\
    \bottomrule
    \end{tabular}
    }
    \caption{Quantitative results on the SUN RGB-D dataset.}
    \label{tab:sun-rgbd}
\end{table}

\begin{table}[t]
    \centering
    \resizebox{0.9\columnwidth}{!}{
    \begin{tabular}{c|ccc|ccc}
    \toprule
       Setting & $\lambda_g$ & $\lambda_l$ & $\lambda_{pred}$ & RMSE $\downarrow$ & Rel $\downarrow$ & $\delta_{1.25}\uparrow$  \\
         \midrule
       A &   - & - & \checkmark & 0.207 & 0.032 & 97.8 \\
       B &   \checkmark & - & \checkmark & 0.212 & 0.038 & 97.8 \\
       C &   - & \checkmark & \checkmark & 0.174 & 0.025 & 98.3 \\
          \cmidrule(lr){1-7}
        D &  0.5 & 1 & 10 & 0.103 & 0.016 & 99.4 \\
         \bottomrule
    \end{tabular}
    }
    \caption{Quantitative comparisons of different $L_1$ loss settings.
     }
    \label{tab:aba_lambda}
\end{table}

\textbf{SUN RGB-D.}
On SUN RGB-D, we adopted the pseudo depth maps as the input and the raw depth data as the ground truth for training.
In the test set, the raw depth image and the depth map synthesized by multiple frames were used as the input and the ground truth, respectively.
In Tab.~\ref{tab:sun-rgbd}, our proposed method achieves the best performance in most metrics.
From the visualization results in Fig.~\ref{fig:sunrgb}, our model complements the missing depth regions as much as possible with more detailed texture information for different sensors.

\subsection{Ablation Studies}
\label{sec:ablation}
We conducted ablation studies with the setting of $\mathcal{T^*} \Rightarrow \mathcal{T}$ on the \mbox{NYU-Depth V2} dataset.

\begin{figure}[t]
\centering
\includegraphics[width=1\linewidth]{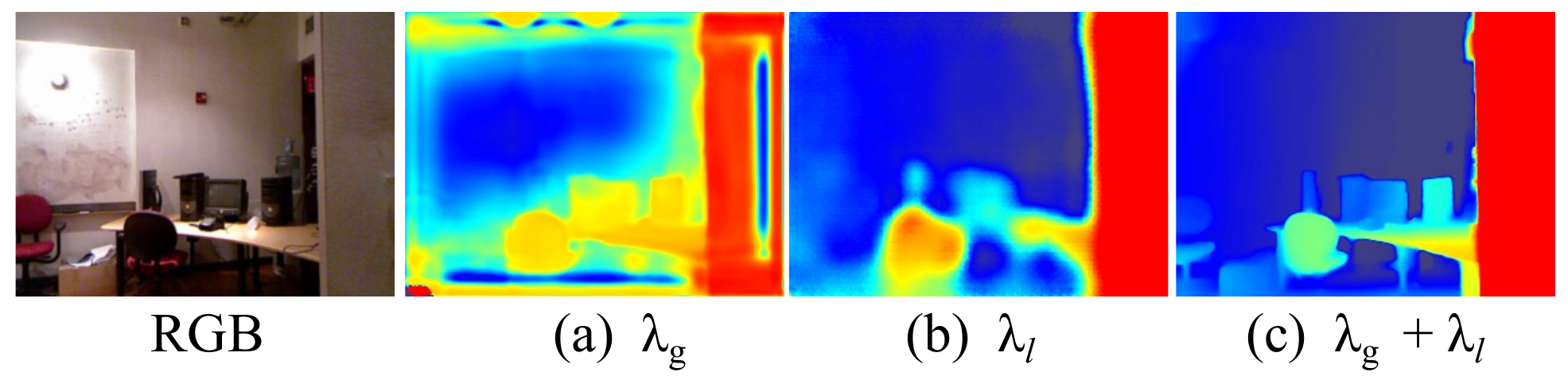}
\caption{Qualitative comparisons of different $L_1$ loss settings.
}
\label{fig:aba_l1}
\end{figure}

\textbf{Settings of $\lambda$s.}
We investigated the effects on model performance in different settings of $\lambda$s in the loss function, and the results are shown in Tab.~\ref{tab:aba_lambda}.
We compared the following four settings and found that including all $L_1$ loss terms leads to the best model.
In Setting A, we only calculated the $L_1$ loss for the final depth prediction, and in Setting~B, both $L_1$ losses for the final depth prediction and fused depth map were calculated.
In these two settings, the model overly focused on textural information resulting in generating many local outliers, as shown in Fig.~\ref{fig:aba_l1}(a), and the predicted depth values in many regions had a large deviation from the ground-truth values.
In Setting C, we took the $L_1$ losses for the local depth map and the final depth prediction.
Although its performance is slightly better, as the model degenerated to the encoder-decoder structure, the depth completion result was shaped towards a blurry depth image, as shown in Fig.~\ref{fig:aba_l1}(b).
Calculating $L_1$ for both branches (Setting~D) is the final setting we adopted, which obtained significant improvement in all metrics and generated the reasonable depth prediction, as shown in Fig.~\ref{fig:aba_l1}(c).

\begin{table}[t]
\centering
\resizebox{0.99\columnwidth}{!}{
    \begin{tabular}{c|c|ccc}
    \toprule
    Module & Method & RMSE $\downarrow$ & REL $\downarrow$ & $\delta_{1.25}\uparrow $ \\
    \midrule
    \multirow{2}{*}{Fusion Head} & Conv. & $0.118$ & $0.022$ & $99.0$  \\
    & Confidence Fusion & $0.117$ & $0.019$ & $99.1$  \\
    \cmidrule(lr){1-5}
    \multirow{4}{*}{Local Guidance}  & Concat. & $0.113$ & $0.017$ & $99.2$  \\
    & U-Net & $0.107$ & $0.016$ & $99.4$  \\
    & U-Net (I) & $0.106$ & $0.016$ & $99.4$  \\
    & U-Net (N) & $0.101$ & $0.015$ & $99.5$  \\
    \cmidrule(lr){1-5}
    \multirow{3}{*}{Stage Fusion}
    & IN & $0.106$ & $0.016$ & $99.4$  \\
    & AdaIN & $0.110$ & $0.017$ & $99.3$  \\
    & {\wada} & $0.103$ & $0.016$ & $99.4$  \\
    \bottomrule
    \end{tabular}
}
\caption{Ablation study results for different modules.
  `Conv.' means the convolution operation for the concatenation of the outputs from the two branches.
  `U-Net (I)' and `U-Net (N)' represent pre-training with ImageNet and NYU-Depth V2, respectively.
  }
\label{tab:aba_module}
\end{table}

\begin{figure}[t]
\centering
\includegraphics[width=.8\linewidth]{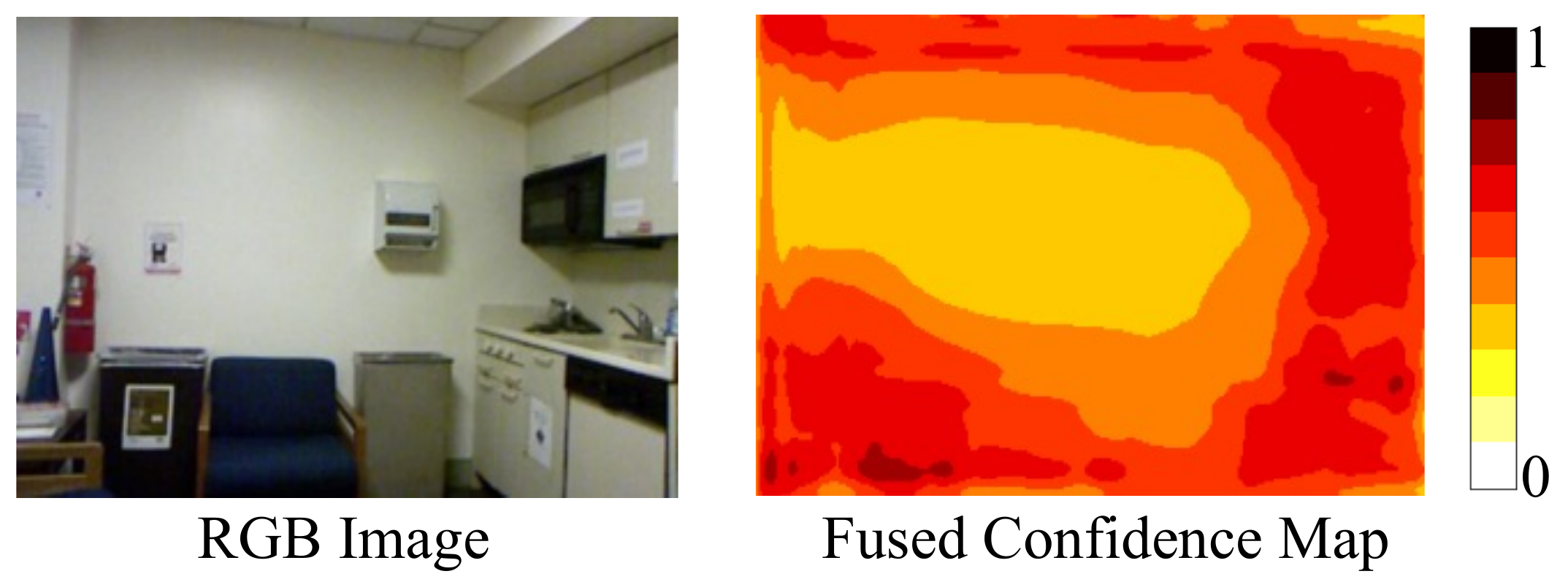}
\caption{Visualization of the confidence map from {\rdfg}.
}

\label{fig:aba_confidence}
\end{figure}

\textbf{Modules.}
On the basis of the two-branch structure, we evaluated the impact of different modules by comparing them with alternative components.
Based on the results shown in Tab.~\ref{tab:aba_module}, we observe the following:

\begin{itemize} [itemsep=1pt,topsep=1pt,parsep=1pt,leftmargin=10pt] 
  \item For the fusion head,
the confidence fusion performs better than the convolution operation~(Conv.).
In addition, Fig.~\ref{fig:aba_confidence} shows the fused confidence map of an RGB image.
The confidence values are high for the foreground objects with richer textural information.
It indicates that RDF-GAN makes better use of rich textural information to improve the depth completion.
  \item Using the local guidance module clearly improves the performance.
The modules using U-Net~\cite{ronneberger2015u} are better than the method of the direct concatenation~(Concat.) of RGB and depth images, and pre-training on ImageNet~\cite{deng2009imagenet} further boosts the performance.
By utilizing additional semantic information of the test scenes, i.e., pre-training U-Net with semantic segmentation on NYU-Depth V2, our method can achieve even better performance.
  \item For the stage fusion modules, {\wada} outperforms the others (IN~\cite{huang2017arbitrary} and AdaIN~\cite{karras2019style}) by a clear margin.
\end{itemize}

\begin{table}[t]
    \centering
    \resizebox{0.7\columnwidth}{!}{
    \begin{tabular}{c|cc}
    \toprule
         Method & mAP@25 & mAP@50 \\
         \midrule
         VoteNet~\cite{qi2019deep} & 59.07 & 35.77 \\
         Ours+VoteNet~\cite{qi2019deep} & \textbf{60.64} & \textbf{37.28} \\
         \cmidrule(lr){1-3}
         H3DNet~\cite{zhang2020h3dnet} & 60.11 & 39.04 \\
         Ours+H3DNet~\cite{zhang2020h3dnet} & \textbf{61.03} & \textbf{39.71} \\
    \bottomrule
    \end{tabular}
    }
    \caption{Performance comparisons of 3D object detection results with the raw and completed depth maps on {SUN RGB-D}.}
    \label{tab:3ddet}
\end{table}

\begin{figure}[t]
\centering
\includegraphics[width=.8\linewidth]{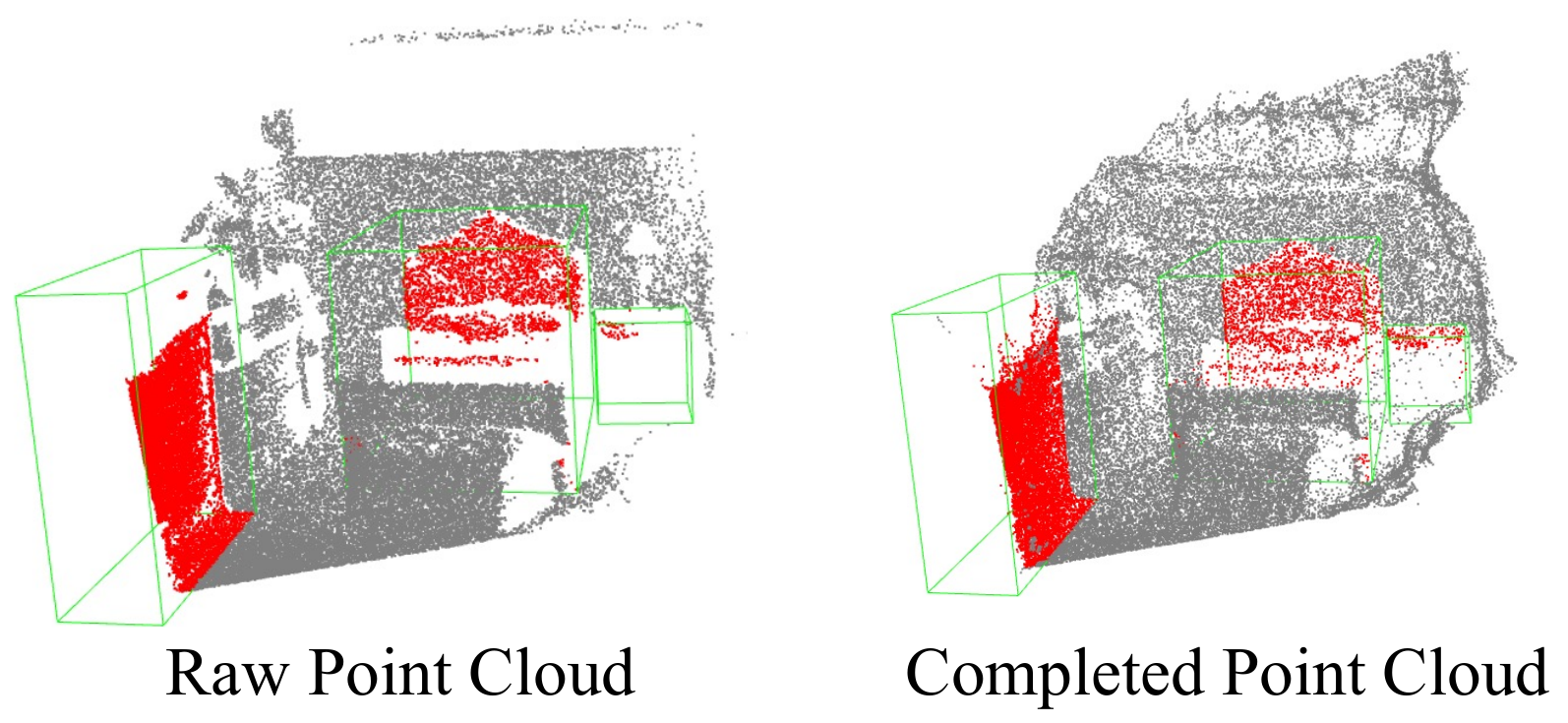}
\caption{Visualizations of point clouds converted by the depth map. The green boxes are the predicted bounding boxes of the detected objects. The red points are the points inside the boxes.}
\label{fig:3ddet}
\end{figure}

\subsection{Object Detection on the Completed Depth Map}
\label{sec:3ddet}
We used the completed depth map as the input of the 3D object detection task on the SUN RGB-D dataset~\cite{song2015sun} to evaluate the quality of our depth completions.
Two SOTA models, VoteNet~\cite{qi2019deep} and H3DNet~\cite{zhang2020h3dnet}, were used as the detectors.
Tab.~\ref{tab:3ddet} shows that the two models both obtain a significant improvement with our completed depth map.
As shown in Fig.~\ref{fig:3ddet}, the point cloud converted from the completed depth map contains more points and better covers the shape of the object than the raw depth map.
More discussions can be found in Section~4 of the supplementary.

\section{Conclusion}
In this work, we propose a novel two-branch end-to-end network for indoor depth completion. 
We design the RDF-GAN model to produce the fine-grained textural depth map and restraint it by a constraint network.
In addition, we propose a novel and effective sampling method to produce pseudo depth maps for training indoor depth completion models. 
Extensive experiments have demonstrated that our proposed solution achieves state-of-the-art on the \mbox{NYU-Depth V2} and \mbox{SUN RGB-D} datasets.

\section{Acknowledgements}
This work was done when Haowen Wang took an internship at Midea Group.
This work was supported in part by the Funds for International Cooperation and Exchange of NSFC under Grant 61720106007, the Fundamental Research Funds for the Central Universities (2019XD-A03-3), the 111 Project under Grant B18008, and Shanghai Pujiang Program (21PJ1420300).

{\small
  \bibliographystyle{ieee_fullname}
  \bibliography{main}

\begin{thebibliography}{10}\itemsep=-1pt

\bibitem{arnold2010automatic}
Mirko Arnold, Anarta Ghosh, Stefan Ameling, and Gerard Lacey.
\newblock Automatic segmentation and inpainting of specular highlights for
  endoscopic imaging.
\newblock {\em EURASIP Journal on Image and Video Processing}, 2010:1--12,
  2010.

\bibitem{xtion}
{ASUS}.
\newblock Asus xtion.
\newblock \url{www.asus.com/Multimedia/Xtion_PRO/}.

\bibitem{cheng2018depth}
Xinjing Cheng, Peng Wang, and Ruigang Yang.
\newblock Depth estimation via affinity learned with convolutional spatial
  propagation network.
\newblock In {\em Proceedings of the European Conference on Computer Vision
  (ECCV)}, pages 103--119, 2018.

\bibitem{cheng2017locality}
Yanhua Cheng, Rui Cai, Zhiwei Li, Xin Zhao, and Kaiqi Huang.
\newblock Locality-sensitive deconvolution networks with gated fusion for rgb-d
  indoor semantic segmentation.
\newblock In {\em Proceedings of the IEEE Conference on Computer Vision and
  Pattern Recognition (CVPR)}, pages 3029--3037, 2017.

\bibitem{deng2009imagenet}
Jia Deng, Wei Dong, Richard Socher, Li-Jia Li, Kai Li, and Li Fei-Fei.
\newblock Imagenet: A large-scale hierarchical image database.
\newblock In {\em Proceedings of the IEEE Conference on Computer Vision and
  Pattern Recognition (CVPR)}, pages 248--255. Ieee, 2009.

\bibitem{du2019translate}
Dapeng Du, Limin Wang, Huiling Wang, Kai Zhao, and Gangshan Wu.
\newblock Translate-to-recognize networks for rgb-d scene recognition.
\newblock In {\em Proceedings of the IEEE/CVF Conference on Computer Vision and
  Pattern Recognition (CVPR)}, June 2019.

\bibitem{felzenszwalb2004efficient}
Pedro~F Felzenszwalb and Daniel~P Huttenlocher.
\newblock Efficient graph-based image segmentation.
\newblock {\em International journal of computer vision (IJCV)},
  59(2):167--181, 2004.

\bibitem{Fu_2018_CVPR}
Yanping Fu, Qingan Yan, Long Yang, Jie Liao, and Chunxia Xiao.
\newblock Texture mapping for 3d reconstruction with rgb-d sensor.
\newblock In {\em Proceedings of the IEEE Conference on Computer Vision and
  Pattern Recognition (CVPR)}, June 2018.

\bibitem{arjovsky2017wasserstein}
Ishaan Gulrajani, Faruk Ahmed, Martin Arjovsky, Vincent Dumoulin, and Aaron~C
  Courville.
\newblock Improved training of wasserstein gans.
\newblock In I. Guyon, U.~V. Luxburg, S. Bengio, H. Wallach, R. Fergus, S.
  Vishwanathan, and R. Garnett, editors, {\em Advances in Neural Information
  Processing Systems}, volume~30. Curran Associates, Inc., 2017.

\bibitem{he2016deep}
Kaiming He, Xiangyu Zhang, Shaoqing Ren, and Jian Sun.
\newblock Deep residual learning for image recognition.
\newblock In {\em Proceedings of the IEEE Conference on Computer Vision and
  Pattern Recognition (CVPR)}, June 2016.

\bibitem{huang2017arbitrary}
Xun Huang and Serge Belongie.
\newblock Arbitrary style transfer in real-time with adaptive instance
  normalization.
\newblock In {\em Proceedings of the IEEE International Conference on Computer
  Vision (ICCV)}, Oct 2017.

\bibitem{huang2019indoor}
Yu-Kai Huang, Tsung-Han Wu, Yueh-Cheng Liu, and Winston~H. Hsu.
\newblock Indoor depth completion with boundary consistency and self-attention.
\newblock In {\em Proceedings of the IEEE/CVF International Conference on
  Computer Vision (ICCV) Workshops}, Oct 2019.

\bibitem{imran2019depth}
Saif Imran, Yunfei Long, Xiaoming Liu, and Daniel Morris.
\newblock Depth coefficients for depth completion.
\newblock In {\em Proceedings of the IEEE/CVF Conference on Computer Vision and
  Pattern Recognition (CVPR)}, June 2019.

\bibitem{isola2017image}
Phillip Isola, Jun-Yan Zhu, Tinghui Zhou, and Alexei~A. Efros.
\newblock Image-to-image translation with conditional adversarial networks.
\newblock In {\em Proceedings of the IEEE Conference on Computer Vision and
  Pattern Recognition (CVPR)}, July 2017.

\bibitem{karras2019style}
Tero Karras, Samuli Laine, and Timo Aila.
\newblock A style-based generator architecture for generative adversarial
  networks.
\newblock In {\em Proceedings of the IEEE/CVF Conference on Computer Vision and
  Pattern Recognition (CVPR)}, June 2019.

\bibitem{Keselman_2017_CVPR_Workshops}
Leonid Keselman, John Iselin~Woodfill, Anders Grunnet-Jepsen, and Achintya
  Bhowmik.
\newblock Intel realsense stereoscopic depth cameras.
\newblock In {\em Proceedings of the IEEE Conference on Computer Vision and
  Pattern Recognition (CVPR) Workshops}, July 2017.

\bibitem{lee2019depth}
Byeong-Uk Lee, Hae-Gon Jeon, Sunghoon Im, and In~So Kweon.
\newblock Depth completion with deep geometry and context guidance.
\newblock In {\em 2019 International Conference on Robotics and Automation
  (ICRA)}, pages 3281--3287. IEEE, 2019.

\bibitem{lee2021depth}
Byeong-Uk Lee, Kyunghyun Lee, and In~So Kweon.
\newblock Depth completion using plane-residual representation.
\newblock In {\em Proceedings of the IEEE/CVF Conference on Computer Vision and
  Pattern Recognition (CVPR)}, pages 13916--13925, June 2021.

\bibitem{li2020multi}
Ang Li, Zejian Yuan, Yonggen Ling, Wanchao Chi, shenghao zhang, and Chong
  Zhang.
\newblock A multi-scale guided cascade hourglass network for depth completion.
\newblock In {\em Proceedings of the IEEE/CVF Winter Conference on Applications
  of Computer Vision (WACV)}, March 2020.

\bibitem{li2019vision}
Bing Li, Juan~Pablo Munoz, Xuejian Rong, Qingtian Chen, Jizhong Xiao, Yingli
  Tian, Aries Arditi, and Mohammed Yousuf.
\newblock Vision-based mobile indoor assistive navigation aid for blind people.
\newblock {\em IEEE Transactions on Mobile Computing (TMC)}, 18(3):702--714,
  2019.

\bibitem{liu2014discrete}
Miaomiao Liu, Mathieu Salzmann, and Xuming He.
\newblock Discrete-continuous depth estimation from a single image.
\newblock In {\em Proceedings of the IEEE Conference on Computer Vision and
  Pattern Recognition (CVPR)}, pages 716--723, 2014.

\bibitem{liu2017learning}
Sifei Liu, Shalini De~Mello, Jinwei Gu, Guangyu Zhong, Ming-Hsuan Yang, and Jan
  Kautz.
\newblock Learning affinity via spatial propagation networks.
\newblock In {\em NIPS}, 2017.

\bibitem{ma2018sparse}
Fangchang Ma and Sertac Karaman.
\newblock Sparse-to-dense: Depth prediction from sparse depth samples and a
  single image.
\newblock In {\em 2018 IEEE international conference on robotics and automation
  (ICRA)}, pages 4796--4803. IEEE, 2018.

\bibitem{ma2019fusiongan}
Jiayi Ma, Wei Yu, Pengwei Liang, Chang Li, and Junjun Jiang.
\newblock Fusiongan: A generative adversarial network for infrared and visible
  image fusion.
\newblock {\em Information Fusion}, 48:11--26, 2019.

\bibitem{maire2016affinity}
Michael Maire, Takuya Narihira, and Stella~X Yu.
\newblock Affinity cnn: Learning pixel-centric pairwise relations for
  figure/ground embedding.
\newblock In {\em Proceedings of the IEEE Conference on Computer Vision and
  Pattern Recognition (CVPR)}, pages 174--182, 2016.

\bibitem{kinect}
{Microsoft}.
\newblock Kinect for windows.
\newblock \url{https://developer.microsoft.com/en-us/windows/kinect/}.

\bibitem{mirza2014conditional}
Mehdi Mirza and Simon Osindero.
\newblock Conditional generative adversarial nets.
\newblock {\em arXiv preprint arXiv:1411.1784}, 2014.

\bibitem{Silberman:ECCV12}
Pushmeet~Kohli Nathan~Silberman, Derek~Hoiem and Rob Fergus.
\newblock Indoor segmentation and support inference from rgbd images.
\newblock In {\em European Conference on Computer Vision (ECCV)}, 2012.

\bibitem{park2020non}
Jinsun Park, Kyungdon Joo, Zhe Hu, Chi-Kuei Liu, and In So~Kweon.
\newblock Non-local spatial propagation network for depth completion.
\newblock In {\em European Conference on Computer Vision (ECCV)}, pages
  120--136. Springer, 2020.

\bibitem{park2017rdfnet}
Seong-Jin Park, Ki-Sang Hong, and Seungyong Lee.
\newblock Rdfnet: Rgb-d multi-level residual feature fusion for indoor semantic
  segmentation.
\newblock In {\em Proceedings of the IEEE International Conference on Computer
  Vision (ICCV)}, pages 4980--4989, 2017.

\bibitem{qi2019deep}
Charles~R Qi, Or Litany, Kaiming He, and Leonidas~J Guibas.
\newblock Deep hough voting for 3d object detection in point clouds.
\newblock In {\em Proceedings of the IEEE International Conference on Computer
  Vision (ICCV)}, pages 9277--9286, 2019.

\bibitem{qiu2019deeplidar}
Jiaxiong Qiu, Zhaopeng Cui, Yinda Zhang, Xingdi Zhang, Shuaicheng Liu, Bing
  Zeng, and Marc Pollefeys.
\newblock Deeplidar: Deep surface normal guided depth prediction for outdoor
  scene from sparse lidar data and single color image.
\newblock In {\em Proceedings of the IEEE Conference on Computer Vision and
  Pattern Recognition (CVPR)}, pages 3313--3322, 2019.

\bibitem{ronneberger2015u}
Olaf Ronneberger, Philipp Fischer, and Thomas Brox.
\newblock U-net: Convolutional networks for biomedical image segmentation.
\newblock In {\em International Conference on Medical image computing and
  computer-assisted intervention}, pages 234--241. Springer, 2015.

\bibitem{saxena2005learning}
Ashutosh Saxena, Sung~H Chung, and Andrew~Y Ng.
\newblock Learning depth from single monocular images.
\newblock In {\em NIPS}, 2005.

\bibitem{senushkin2020decoder}
Dmitry Senushkin, Mikhail Romanov, Ilia Belikov, Nikolay Patakin, and Anton
  Konushin.
\newblock Decoder modulation for indoor depth completion.
\newblock In {\em {IEEE/RSJ} International Conference on Intelligent Robots and
  Systems, {IROS} 2021, Prague, Czech Republic, September 27 - Oct. 1, 2021},
  pages 2181--2188. {IEEE}, 2021.

\bibitem{song2015sun}
Shuran Song, Samuel~P Lichtenberg, and Jianxiong Xiao.
\newblock Sun rgb-d: A rgb-d scene understanding benchmark suite.
\newblock In {\em Proceedings of the IEEE Conference on Computer Vision and
  Pattern Recognition (CVPR)}, pages 567--576, 2015.

\bibitem{van2019sparse}
Wouter Van~Gansbeke, Davy Neven, Bert De~Brabandere, and Luc Van~Gool.
\newblock Sparse and noisy lidar completion with rgb guidance and uncertainty.
\newblock In {\em 2019 16th international conference on machine vision
  applications (MVA)}, pages 1--6. IEEE, 2019.

\bibitem{vaswani2017attention}
Ashish Vaswani, Noam Shazeer, Niki Parmar, Jakob Uszkoreit, Llion Jones,
  Aidan~N Gomez, {\L}ukasz Kaiser, and Illia Polosukhin.
\newblock Attention is all you need.
\newblock In {\em NIPS}, pages 5998--6008, 2017.

\bibitem{yang2014stereo}
Qingxiong Yang.
\newblock Stereo matching using tree filtering.
\newblock {\em IEEE Transactions on Pattern Analysis and Machine Intelligence
  (PAMI)}, 37(4):834--846, 2014.

\bibitem{depthcompletion_gan}
Chongzhen Zhang, Yang Tang, Chaoqiang Zhao, Qiyu Sun, Zhencheng Ye, and Jürgen
  Kurths.
\newblock Multitask gans for semantic segmentation and depth completion with
  cycle consistency.
\newblock {\em IEEE Transactions on Neural Networks and Learning Systems},
  32(12):5404--5415, 2021.

\bibitem{zhang2020h3dnet}
Zaiwei Zhang, Bo Sun, Haitao Yang, and Qixing Huang.
\newblock H3dnet: 3d object detection using hybrid geometric primitives.
\newblock In {\em Proceedings of the European Conference on Computer Vision
  (ECCV)}, pages 311--329. Springer, 2020.

\bibitem{zhao2020pointar}
Yiqin Zhao and Tian Guo.
\newblock Pointar: Efficient lighting estimation for mobile augmented reality.
\newblock In {\em European Conference on Computer Vision (ECCV)}, pages
  678--693. Springer, 2020.

\end{thebibliography}


\begin{thebibliography}{10}\itemsep=-1pt

\bibitem{arnold2010automatic}
Mirko Arnold, Anarta Ghosh, Stefan Ameling, and Gerard Lacey.
\newblock Automatic segmentation and inpainting of specular highlights for
  endoscopic imaging.
\newblock {\em EURASIP Journal on Image and Video Processing}, 2010:1--12,
  2010.

\bibitem{baek2020distance}
Eu-Tteum Baek, Hyung-Jeong Yang, Soo-Hyung Kim, Gueesang Lee, and Hieyong
  Jeong.
\newblock Distance error correction in time-of-flight cameras using
  asynchronous integration time.
\newblock {\em Sensors}, 20(4):1156, 2020.

\bibitem{cheng2018depth}
Xinjing Cheng, Peng Wang, and Ruigang Yang.
\newblock Depth estimation via affinity learned with convolutional spatial
  propagation network.
\newblock In {\em Proceedings of the European Conference on Computer Vision
  (ECCV)}, pages 103--119, 2018.

\bibitem{felzenszwalb2004efficient}
Pedro~F Felzenszwalb and Daniel~P Huttenlocher.
\newblock Efficient graph-based image segmentation.
\newblock {\em International journal of computer vision (IJCV)},
  59(2):167--181, 2004.

\bibitem{imran2019depth}
Saif Imran, Yunfei Long, Xiaoming Liu, and Daniel Morris.
\newblock Depth coefficients for depth completion.
\newblock In {\em Proceedings of the IEEE/CVF Conference on Computer Vision and
  Pattern Recognition (CVPR)}, June 2019.

\bibitem{kim2013depth}
Sung-Yeol Kim, Manbae Kim, and Yo-Sung Ho.
\newblock Depth image filter for mixed and noisy pixel removal in rgb-d camera
  systems.
\newblock {\em IEEE Transactions on Consumer Electronics}, 59(3):681--689,
  2013.

\bibitem{lee2019depth}
Byeong-Uk Lee, Hae-Gon Jeon, Sunghoon Im, and In~So Kweon.
\newblock Depth completion with deep geometry and context guidance.
\newblock In {\em 2019 International Conference on Robotics and Automation
  (ICRA)}, pages 3281--3287. IEEE, 2019.

\bibitem{lee2021depth}
Byeong-Uk Lee, Kyunghyun Lee, and In~So Kweon.
\newblock Depth completion using plane-residual representation.
\newblock In {\em Proceedings of the IEEE/CVF Conference on Computer Vision and
  Pattern Recognition (CVPR)}, pages 13916--13925, June 2021.

\bibitem{ma2018sparse}
Fangchang Ma and Sertac Karaman.
\newblock Sparse-to-dense: Depth prediction from sparse depth samples and a
  single image.
\newblock In {\em 2018 IEEE international conference on robotics and automation
  (ICRA)}, pages 4796--4803. IEEE, 2018.

\bibitem{park2020non}
Jinsun Park, Kyungdon Joo, Zhe Hu, Chi-Kuei Liu, and In So~Kweon.
\newblock Non-local spatial propagation network for depth completion.
\newblock In {\em European Conference on Computer Vision (ECCV)}, pages
  120--136. Springer, 2020.

\bibitem{qi2019deep}
Charles~R Qi, Or Litany, Kaiming He, and Leonidas~J Guibas.
\newblock Deep hough voting for 3d object detection in point clouds.
\newblock In {\em Proceedings of the IEEE International Conference on Computer
  Vision (ICCV)}, pages 9277--9286, 2019.

\bibitem{qiu2019deeplidar}
Jiaxiong Qiu, Zhaopeng Cui, Yinda Zhang, Xingdi Zhang, Shuaicheng Liu, Bing
  Zeng, and Marc Pollefeys.
\newblock Deeplidar: Deep surface normal guided depth prediction for outdoor
  scene from sparse lidar data and single color image.
\newblock In {\em Proceedings of the IEEE Conference on Computer Vision and
  Pattern Recognition (CVPR)}, pages 3313--3322, 2019.

\bibitem{ronneberger2015u}
Olaf Ronneberger, Philipp Fischer, and Thomas Brox.
\newblock U-net: Convolutional networks for biomedical image segmentation.
\newblock In {\em International Conference on Medical image computing and
  computer-assisted intervention}, pages 234--241. Springer, 2015.

\bibitem{senushkin2020decoder}
Dmitry Senushkin, Mikhail Romanov, Ilia Belikov, Nikolay Patakin, and Anton
  Konushin.
\newblock Decoder modulation for indoor depth completion.
\newblock In {\em {IEEE/RSJ} International Conference on Intelligent Robots and
  Systems, {IROS} 2021, Prague, Czech Republic, September 27 - Oct. 1, 2021},
  pages 2181--2188. {IEEE}, 2021.

\bibitem{zhang2020h3dnet}
Zaiwei Zhang, Bo Sun, Haitao Yang, and Qixing Huang.
\newblock H3dnet: 3d object detection using hybrid geometric primitives.
\newblock In {\em Proceedings of the European Conference on Computer Vision
  (ECCV)}, pages 311--329. Springer, 2020.

\end{thebibliography}
}

\end{document}


\title{Supplementary Materials for\\RGB-Depth Fusion GAN for Indoor Depth Completion}


\maketitle


\section{Regular Downsampled Input vs. Raw input}
\begin{figure}[h]
    \centering
    \vspace{-10pt}
    \includegraphics[width=1\linewidth]{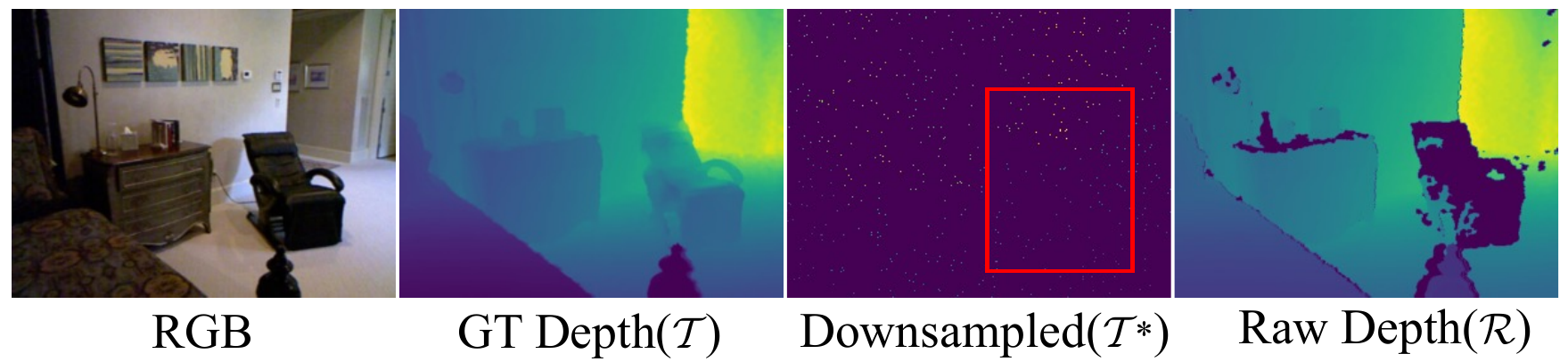}\\
    \includegraphics[width=1\linewidth]{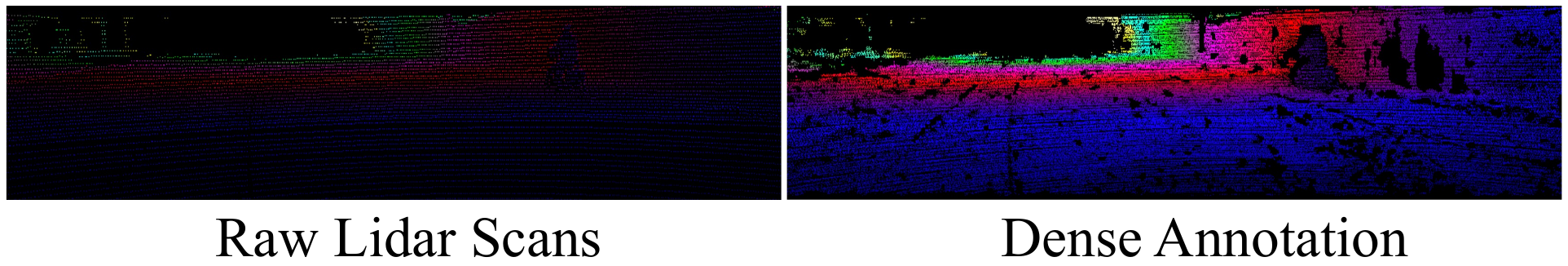}
    \vspace{-18pt}
    \caption{Depth data visualizations of indoor RGB-Depth sensors (top, NYU-Depth~V2) and outdoor Lidar scans (bottom, KITTI).}
    \vspace{-10pt}
    \label{fig:setting}
\end{figure}

The regular downsampled setting of most existing methods following Ma and Karaman~\cite{ma2018sparse} mimics well the task of outdoor depth completion from \textit{raw Lidar scans} to \textit{dense annotations}, as shown in the bottom of Fig.~\ref{fig:setting}.
However, for indoor RGB-depth sensor data, directly using downsampled input is improper:
1) The \textit{raw depth} $\mathcal{R}$ captured by depth sensors is dense and continuous, which is quite different from the sparse pattern of \textit{downsampled input} $\mathcal{T^*}$;
2) As shown in the red box in Fig.~\ref{fig:setting}, the downsampled input reveals ground truth depth values to the models that can not be obtained in practice.
Thus, we believe the raw input setting ($\mathcal{R} \Rightarrow \mathcal{T}$) is more practicable (and not only a specific case) for indoor depth completion than $\mathcal{T^*} \Rightarrow \mathcal{T}$.

\section{More Details of Pseudo Depth Maps}

Section~3.5 of the main paper introduces our proposed pseudo depth maps for training indoor depth completion methods.
In this section, we provide more details about the design for the pseudo depth maps, including how and why the five masking methods are used for generating pseudo depth maps and a few more visualization results.

\begin{figure*}[t]
  \centering
  \includegraphics[width=1\linewidth]{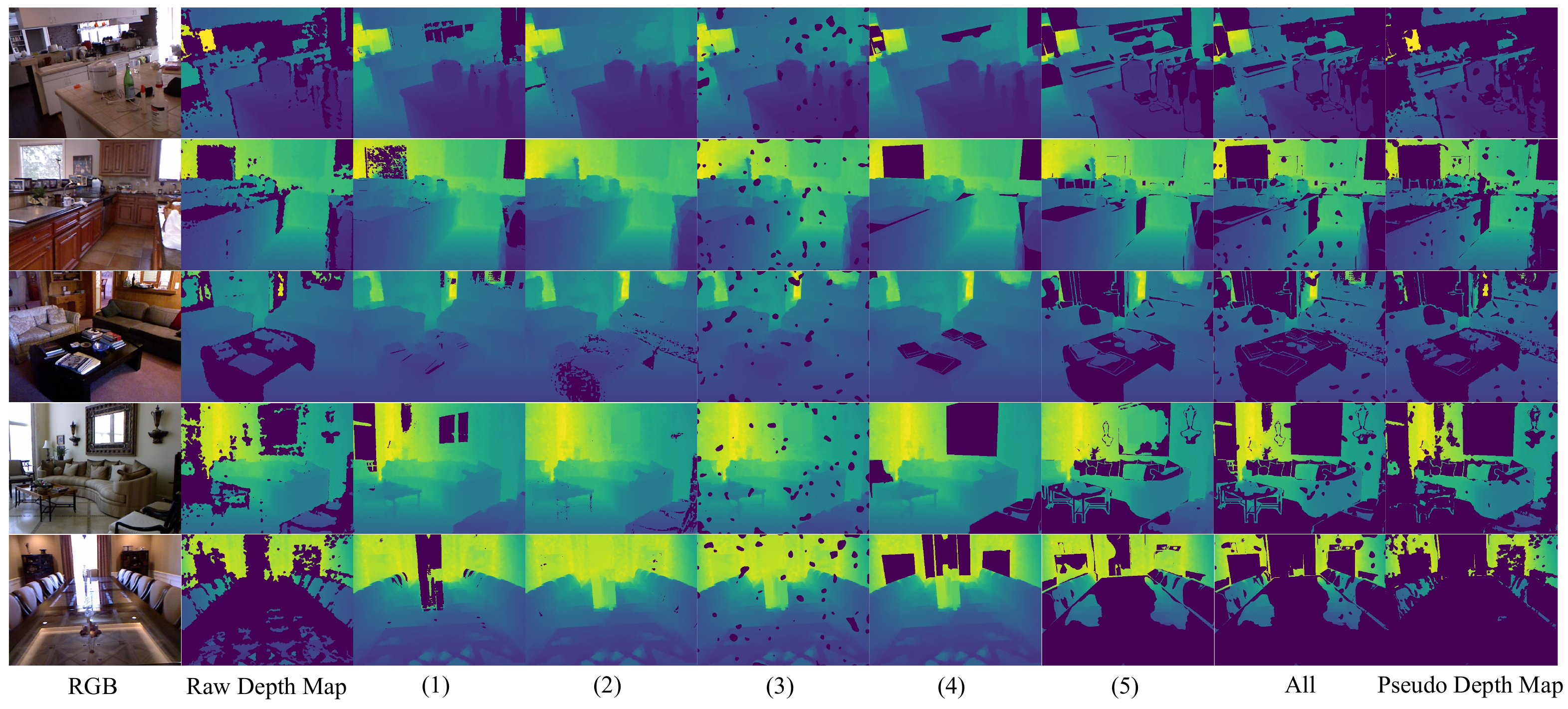}
  \caption{Visualizations of the five sampling methods.
  (1) Highlight masking. (2) Black masking. (3) Graph-based segmentation masking. (4) Semantic masking. (5) Semantic XOR masking.
  `All' refers to the results of using all methods on the reconstruction of the depth map.
  `Pseudo Depth Map' refers to the results of using all methods on the raw depth map.}
  \label{fig:pseudo}
\end{figure*}

\begin{enumerate} [itemsep=1pt,topsep=1pt,parsep=1pt,leftmargin=20pt,label=(\arabic*)] 
\item \textit{Highlight masking}.
The RGB-D camera has difficulty in obtaining depth data of shiny surfaces because IR rays reflected from these surfaces are weak or scattered~\cite{kim2013depth}.
Meanwhile, these smooth and shiny objects often lead to specular highlights and bright spots in the RGB images.
Hence, we detect these highlight regions in RGB images and mask them in depth maps to generate pseudo depth maps.
We borrow from Arnold \etal~{\cite{arnold2010automatic}} for highlight detection, which has a good balance of effectiveness and efficiency.

\item \textit{Black masking}.
Since dark and matte surfaces are good absorbers and poor reflectors of radiation, the depth map is strongly affected by these surfaces~\cite{baek2020distance}.
We randomly mask the depth pixels whose values of R, G, and B in the RGB images are all in [0, 5], which can simply but directly handle some regions that are easy to have invalid depth values.

\item \textit{Graph-based segmentation masking}.
The chaotic light reflections in the complex environment can interfere with the return of infrared light and cause discrete and irregular noises in depth maps.
We use the graph-based segmentation~\cite{felzenszwalb2004efficient} to divide the RGB image into several blocks of different sizes and mask the small blocks.

\item \textit{Semantic masking}.
Some materials, such as glass, mirror, and porcelain surfaces, easily cause scattered infrared reflection and missing depth return values.
We utilize the semantic label information to randomly cover objects probably containing these materials, such as TV, mirror, and window.
We randomly mask all pixels for one or two objects in each frame.

\item \textit{Semantic XOR masking}.
Similar motivations to the graph-based segmentation masking, we use semantic segmentation to recognize complex regions in the scene.
We use the U-Net~\cite{ronneberger2015u} network to randomly partition 20\% of the training set for semantic segmentation task training and subsequently use it to semantically segment the remaining data.
We take the regions where the predicted segmentation results are different from the ground-truth to be the complex regions, then mask the depth values in those regions.

\end{enumerate}

Fig.~\ref{fig:pseudo} shows the quantitative results for each downsampling method.
In Fig.~\ref{fig:pseudo}(1), the highlight regions we masked are basically the depth missing regions of the raw depth images.
We only randomly mask some sporadic black areas since the RGB image has a certain deviation from the real color, in Fig.~\ref{fig:pseudo}(2). Graph-based segmentation masking simulates some discrete depth loss very well of depth maps in Fig.~\ref{fig:pseudo}(3). In Fig.~\ref{fig:pseudo}(4), semantic masking covers out some objects that may cause a lack of depth values. Semantic XOR masking masks a wide range of regions where the predicted and ground-truth values differ in Fig.~\ref{fig:pseudo}(5).

\section{Three Training and Evaluation Settings}
\begin{table}[h]
    \centering
    \resizebox{0.95\columnwidth}{!}{
    \begin{tabular}{c|cc|ccc}
        \toprule
        \multirow{2}{*}{Setting} & \multicolumn{2}{|c|}{Training} & \multicolumn{3}{|c}{Testing} \\
        & {$\mathcal{P} \Rightarrow \mathcal{R}$} & {$\mathcal{R^*} \Rightarrow \mathcal{R}$} & {$\mathcal{R} \Rightarrow \mathcal{T}$} & {$\mathcal{R^*} \Rightarrow \mathcal{T}$} & {$\mathcal{T^*} \Rightarrow \mathcal{T}$}  \\
        \midrule
        A & \checkmark & & \checkmark & & \\
        B & & \checkmark & & \checkmark &  \\
        C & & \checkmark & & & \checkmark \\
        \bottomrule
    \end{tabular}
    }
    \caption{$\mathcal{R}$, $\mathcal{T}$ and $\mathcal{P}$ represent the raw, reconstructed, and pseudo depth map, respectively. $\cdot^*$ represents the random sparse sampling with 500 valid depth pixels.}
    \label{tab:sota_setting}
\end{table}

In the main paper, we set up three different test methods and corresponding training strategies, as shown in Tab.~\ref{tab:sota_setting}. {$\mathcal{R}$} and {$\mathcal{T}$} represent raw or incompleted depth images and reconstructed and completed depth maps, respectively. In the training set, due to the deficiency of a large number of reconstructed depth maps, most methods downsample the raw or incompleted depth images to predict the valid pixels of raw depth maps. In our work, we use the pseudo depth maps for training. In addition, we randomly sample 500 valid points {$\mathcal{R^*}$} to get the sparse depth map as the input following existing methods~\cite{ma2018sparse, lee2021depth}. The specific three evaluation programs are set up as follows:

\begin{itemize} [itemsep=1pt,topsep=1pt,parsep=1pt,leftmargin=10pt] 
\item \textit{Setting A}: At the training time, we use pseudo depth maps {$\mathcal{P}$} as model input, and supervise with raw depth image. In testing, we input a raw depth map to predict the complemented and reconstructed depth map, which is most in line with the real scenario of indoor depth completion. Our method uses the pseudo depth maps, and other methods are trained in the synthetic semi-dense sensor data~\cite{senushkin2020decoder}.

\item \textit{Setting B}: Although our model is not designed for sparse scenes, we use the sparse depth map {$\mathcal{R^*}$} with randomly sampled 500 valid depth pixels following existing methods~\cite{ma2018sparse,cheng2018depth,lee2021depth} for training to evaluate the model completion performance.  At the test time, the input is consistent with the sampling method of training for raw depth images, and the reconstructed depth map is used as the ground truth for evaluation.

\item \textit{Setting C}: For comparing more existing methods~\cite{ma2018sparse,imran2019depth,lee2019depth,cheng2018depth,qiu2019deeplidar,park2020non,lee2021depth} of depth completion, we randomly sample the 500 pixels in the reconstructed depth map as input at the test phase. This sampling method, despite the fact that only 500 valid points are the input, would have much better metrics than the above two sampling methods because of the accurate depth information obtained for all regions.

\end{itemize}

\section{Object Detection after Depth Completion}
\begin{table}[h]
    \centering
    \resizebox{1\columnwidth}{!}{
    \begin{tabular}{c|cc|c}
    \toprule
         Method & mAP@25 & mAP@50 & RMSE\\
         \midrule
         VoteNet~\cite{qi2019deep} & 59.07 & 35.77 & -\\
         DeepLidar~\cite{qiu2019deeplidar} + VoteNet~\cite{qi2019deep} & 59.73 & 35.49 & 0.279\\
         NLSPN~\cite{park2020non} + VoteNet~\cite{qi2019deep} & 47.43 & 26.10 & 0.267\\
         Ours + VoteNet~\cite{qi2019deep} & \textbf{60.64} & \textbf{37.28} & \textbf{0.255}\\
         \cmidrule(lr){1-4}
         H3DNet~\cite{zhang2020h3dnet} & 60.11 & 39.04 & -\\
         DeepLidar~\cite{qiu2019deeplidar} + H3DNet~\cite{zhang2020h3dnet} & 60.35 & 39.16 & 0.279\\
         NLSPN~\cite{park2020non} + H3DNet~\cite{zhang2020h3dnet} & 27.10 & 9.77 & 0.267\\
         Ours + H3DNet~\cite{zhang2020h3dnet} & \textbf{61.03} & \textbf{39.71} & \textbf{0.255}\\
    \bottomrule
    \end{tabular}
    }
    \caption{Comparisons of 3D object detection results with the completed depth map on {SUN RGB-D}. The last column is the complementary result for DeepLidar, NLSPN, and Ours.}
    \label{tab:3ddet}
\end{table}

We show extended experimental results using completed depth maps for 3D object detection, of which some representative results are shown in Section~4.4 in the main paper.
We compare with the depth maps generated by DeepLidar~\cite{qiu2019deeplidar} and NLSPN~\cite{park2020non} on the 3D object detection task. DeepLidar~\cite{qiu2019deeplidar} uses a surface normal pathway to assist in depth map completion.
NLSPN~\cite{park2020non} learns the convolutional kernel size and iteration number for propagation to optimize the boundary depth. 
In Tab.~\ref{tab:3ddet}, compared to DeepLidar~\cite{qiu2019deeplidar}, our model improves more significantly in all metrics.
NLSPN~\cite{park2020non} produces too much noise in the completion, which causes the performance of the detector to degrade.

{\small
  \bibliographystyle{ieee_fullname}
\bibliography{supp}
}